\documentclass[sigconf,screen]{acmart}
\AtBeginDocument{%
  }

\setcopyright{acmlicensed}
\copyrightyear{2025} 
\acmYear{2025} 
\setcopyright{cc}
\setcctype{by}
\acmConference[MM '25]{Proceedings of the 33rd ACM International Conference
on Multimedia}{October 27--31, 2025}{Dublin, Ireland}
\acmBooktitle{Proceedings of the 33rd ACM International Conference on
Multimedia (MM '25), October 27--31, 2025, Dublin,
Ireland}\acmDOI{10.1145/3746027.3755010}
\acmISBN{979-8-4007-2035-2/2025/10}

\acmSubmissionID{}

\usepackage{listings}
\usepackage[title]{appendix}
\definecolor{codegreen}{rgb}{0,0.6,0}
\definecolor{codegray}{rgb}{0.5,0.5,0.5}
\definecolor{codepurple}{rgb}{0.58,0,0.82}
\definecolor{backcolour}{rgb}{0.95,0.95,0.92}

\lstdefinestyle{mystyle}{
    backgroundcolor=\color{backcolour},   
    commentstyle=\color{codegreen},
    keywordstyle=\color{magenta},
    numberstyle=\tiny\color{codegray},
    stringstyle=\color{codepurple},
    basicstyle=\ttfamily\footnotesize,
    breakatwhitespace=false,         
    breaklines=true,                 
    captionpos=b,                    
    keepspaces=true,                 
    numbersep=5pt,                  
    showspaces=false,                
    showstringspaces=false,
    showtabs=false,                  
    tabsize=2
}
\definecolor{gray}{rgb}{0.5,0.5, 0.5}
\definecolor{third}{rgb}{1.0000, 0.9355, 0.7843} 

\definecolor{second}{rgb}{1.0000, 0.8157, 0.6078} 
\definecolor{best}{rgb}{1.0000, 0.6902, 0.6902} 

\definecolor{s_color}{RGB}{31, 144, 224} 
\definecolor{a_color}{RGB}{204, 57, 189} 

\usepackage{colortbl}
\usepackage{adjustbox}
\usepackage{diagbox}
\usepackage{pifont}

\usepackage{xcolor}
\usepackage{multirow}

\newcommand{\cmark}{\textcolor[rgb]{0.0, 0.5, 0.0}{\ding{51}}}
\newcommand{\xmark}{\textcolor{red}{\ding{55}}}    

\begin{document}

\title{DiffArtist: Towards Structure and Appearance Controllable Image Stylization}
\author{Ruixiang Jiang}
\orcid{0000-0001-8666-6767}
\email{rui-x.jiang@connect.polyu.hk}
\affiliation{%
  \institution{The Hong Kong Polytechnic University}
  \city{Hong Kong}
  \country{China}
}

\author{Chang Wen Chen}
\orcid{0000-0002-6720-234X}
\email{chen.changwen@polyu.edu.hk}
\affiliation{%
  \institution{The Hong Kong Polytechnic University}
  \city{Hong Kong}
  \country{China}
}

\renewcommand{\shortauthors}{Ruixiang Jiang and Chang Wen Chen}

\begin{abstract}
Artistic styles are defined by both their structural and appearance elements. Existing neural stylization techniques primarily focus on transferring appearance-level features such as color and texture, often neglecting the equally crucial aspect of structural stylization. To address this gap, we introduce \textbf{DiffArtist}, the first 2D stylization method to offer fine-grained, simultaneous control over both structure and appearance style strength. This dual controllability is achieved by representing structure and appearance generation as separate diffusion processes, necessitating no further tuning or additional adapters. To properly evaluate this new capability of dual stylization, we further propose a Multimodal LLM-based stylization evaluator that aligns significantly better with human preferences than existing metrics. Extensive analysis shows that DiffArtist achieves superior style fidelity and dual-controllability compared to state-of-the-art methods. Its text-driven, training-free design and unprecedented dual controllability make it a powerful and interactive tool for various creative applications.  Project homepage: \url{https://diffusionartist.github.io}.
\end{abstract}



\maketitle

\section{Introduction}

\begin{figure}[!t]
    \centering
    \includegraphics[width=\linewidth]{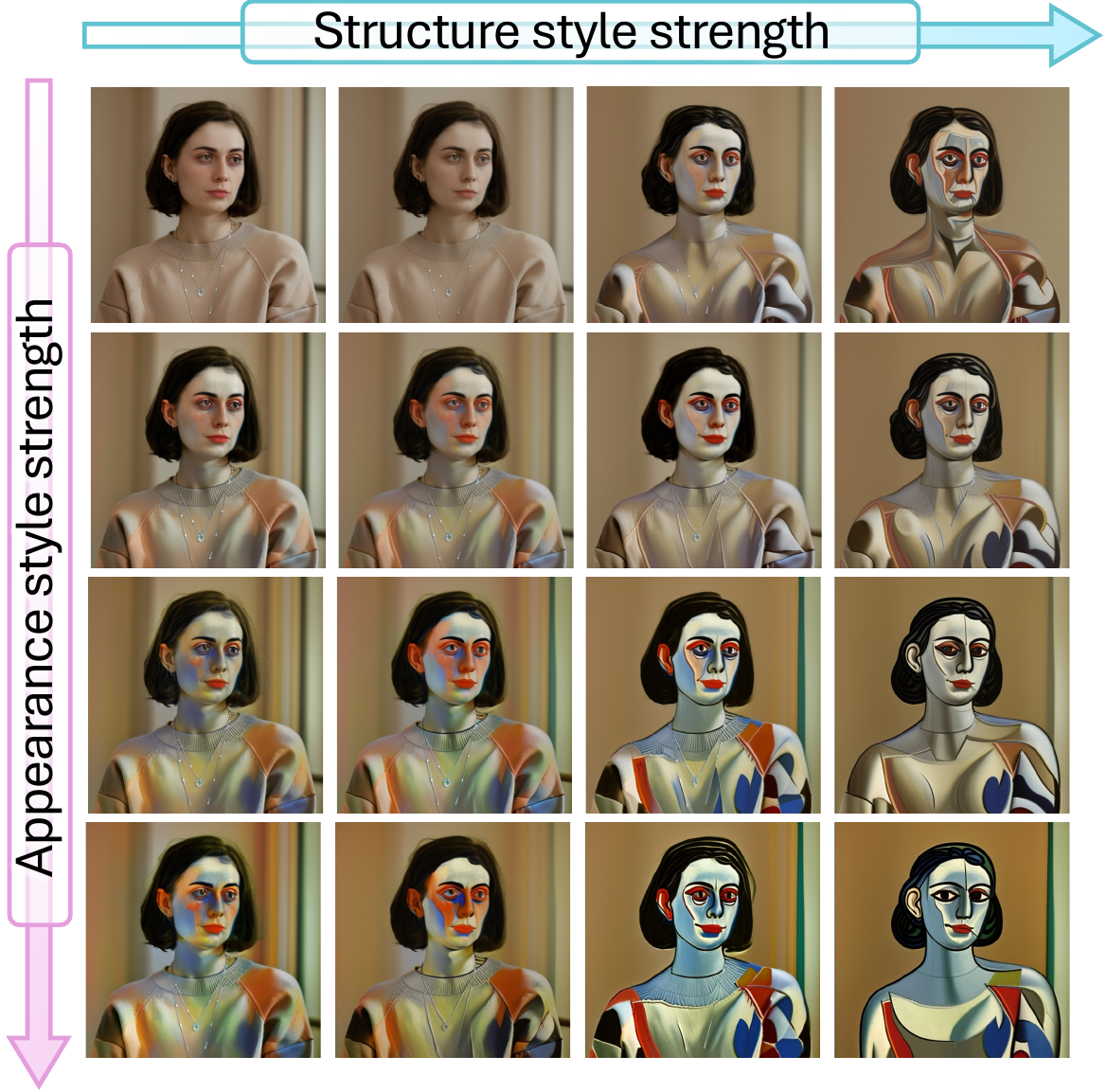}
    \caption{DiffArtist enables disentangled and fine-grained control of style strength from two orthogonal perspectives: structure and appearance. The style prompt is \textit{``The Dream, by Picasso.''}}
    \label{fig:teaser}
\end{figure}

The essence of an artistic style lies not only in its appearance—color and texture—but also its structure—geometry and composition~\cite{gombrich1995story,jung2024geometry,goodman1976languages}. For example, the fragmentation of figures in Picasso's Cubist works and the undulating sky in Van Gogh's ``Starry Night'', each contributing distinctly to their artistic expression. 
Existing neural stylization approaches~\cite{huang2017arbitrary, deng2022stytr2, kwon2022clipstyler, wang2022clast, huang2024diffstyler, cui2023instastyle} predominantly focus on manipulating appearance-level attributes. The structural elements in the source image, however, are often viewed as part of ``content'' and are explicitly preserved~\cite{wang2024instantstylep,wen2023cap,yang2022industrial,yang2023zero}. This fundamental limitation prevents them from capturing the true essence of an art style, severely restricting their expressive potential and customizability.

The root of this limitation lies in the inherent complexity of structural stylization. Unlike appearance-style transfer, structural stylization requires a delicate balance between three competing objectives: (1) aligning with the target style, (2) harmonizing with the source image's composition,  and (3) preserving the core semantic integrity of the content. These objectives operate at a high-level semantic plane, exposing a critical gap in current methods. While descriptors like AdaIN~\cite{huang2017arbitrary} and Gram loss~\cite{gatys2015neural} may suffice for appearance-style modeling, the lack of adequate structure representation and structure-style evaluators presents significant obstacles in the development of structural stylization techniques. This challenge is amplified in recent multimodal generation scenarios, where a style prompt offers no explicit visual template~\cite{huang2024diffstyler,wang2023nerf,cui2023instastyle,hertz2024style}.

The advent of Diffusion Models (DMs) offers a powerful new paradigm for achieving this dual controllability, as their generative sampling process enables far greater structural and appearance diversity than prior methods. This reframes stylization as a conditional generation task, guided by a source image and a style prompt (image or text).  However, this generative power comes with a critical, unaddressed challenge: the diffusion process inherently \textbf{entangles} the generation of structure and appearance. We identify this as a fundamental \textbf{Structure-Appearance (S-A) Tradeoff}: intensifying structural changes inadvertently corrupts appearance style, while strengthening appearance washes out structural transformations. This tradeoff directly explains the core failures of existing diffusion-based methods, which are either prone to severe content degradation~\cite{chung2024style,lyu2023infostyler,wang2024instantstylep} or suffer from weak, constrained stylization~\cite{wang2023stylediffusion,yang2023zero}. Achieving dual controllability in the stylization thus remains an open question.

To solve this, we introduce \textbf{DiffArtist}, the first framework to our knowledge that offers explicit, disentangled control over both structure and appearance in 2D stylization. At its core, DiffArtist explicitly disentangles the structural and appearance generation as separated diffusion processes, with shared semantic information. This design directly overcomes the fundamental S-A tradeoff and functions as a zero-shot, plug-and-play module for any pretrained U-Net-based DM, requiring no costly fine-tuning or external adapters~\cite{zhang2023adding,ye2023ip}. As evidenced in this paper, this design provides true disentanglement—a key advantage over ControlNet-based methods~\cite{wang2024instantstyle,lin2024ctrl} where adjusting one style factor adversely impacts the other. As demonstrated in Fig.~\ref{fig:teaser}, this unprecedented level of control allows DiffArtist to achieve strong, semantically coherent stylization, unleashing the full creative potential of dual-style customization.

Evaluating this novel capability of dual control requires understanding on image semantics, where existing evaluation metrics obsolete. This exposes a critical need for a new evaluation paradigm. To address this, we introduce our second major contribution: \textbf{a Multimodal LLM (MLLM)-based evaluator} designed for dual stylization. We argue that any such evaluator must satisfy three key criteria: (1) operate at a \textbf{high-level semantic plane} to assess structure, (2) possess \textbf{contextual awareness} to maintain semantic integrity, and (3) perform robust \textbf{cross-modal association} between text prompts and visual forms. By leveraging the zero-shot reasoning of MLLMs, our proposed metric meets these criteria. We empirically show that it aligns significantly better with human artistic judgment than existing stylization metrics~\cite{kirstain2023pick,wang2004image,radford2021learning}, establishing a more reliable and human-centric standard for future stylization research.

We summarize our contributions as follows:
\begin{enumerate}
\item We identify the S-A tradeoff in diffusion models as the key challenge for disentangled dual controllability.
\item We propose DiffArtist, the first 2D stylization method that enables the dual controllability of structure and appearance.
\item We present a novel MLLM-based evaluator for evaluating structure and appearance in artistic stylization, which aligns better with human perception.
\item Extensive experiments demonstrate that DiffArtist achieves superior stylization fidelity, control editability, and disentanglement than existing approaches.
\end{enumerate}
\section{Related Works}

\begin{figure*}[!t]
    \centering
    \includegraphics[width=\linewidth]{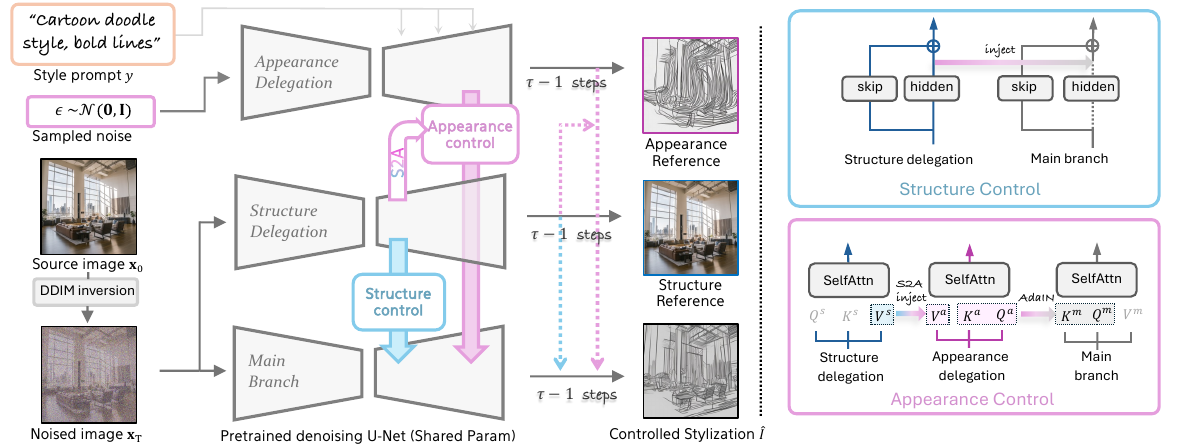}
    \caption{\textbf{Overview of DiffArtist}.Our method disentangles stylization by processing structure and appearance through two independent diffusion trajectories (delegations). At each denoising step, the main stylization branch is conditioned on semantic-level features from the structure and appearance delegations. All three branches share the same pretrained U-Net parameters, and perform full denoising of $\tau$ steps. The entire framework operates without requiring any fine-tuning or adapters.}
    \label{fig:overview}
\end{figure*}

\textbf{The Stylization of Structure and/or Appearance.} Structure and appearance collectively define the style of a visual representation. Existing neural stylization methods~\cite{huang2017arbitrary, deng2022stytr2, kwon2022clipstyler, wang2022clast, huang2024diffstyler, cui2023instastyle} have predominantly focused on appearance stylization, generally achieved via an encoder-decoder architecture.  Only a few papers~\cite{kim2020deformable,zhou2024deformable} focus on transferring structural style components between 2D images. This is usually accomplished by calculating a correspondence between content and style image and performing a non-rigid deformation. However, these methods typically operate on images in specific domains like portraits, requiring an in-domain stylized reference. Moreover, they do not enable dual controllability. Very recently, dual controllability has been explored in Ctrl-X~\cite{lin2024ctrl}, but its framework is designed for localized, image edits rather than the global, harmonious style transformations.  Meanwhile, the concept of disentangled control is explored in 3D synthesis, where explicit representations of shape (e.g., meshes~\cite{michel2022text2mesh} and radiance fields~\cite{ wang2023nerf}) and appearance (e.g., texture mapping~\cite{chen2023text2tex}) make separation natural. The success in 3D, however, is not directly transferable to the 2D domain due to the lack of explicit geometric priors in single images. In this paper, we explore the first dual controllable 2D image stylization method.

\textbf{Text-Driven Image Stylization.}
Text-driven image stylization aims to stylize a source image according to style prompts. Early methods achieve it by optimizing certain image representation~\cite{michel2022text2mesh, kwon2022clipstyler, wang2023nerf, gal2022stylegan, kim2022diffusionclip} with a multimodal alignment objective, typically implemented as the CLIP loss~\cite{radford2021learning}. Recently, it was discovered that text-to-image (T2I) DMs could also be adapted for similar optimization schemes~\cite{poole2022dreamfusion, hertz2023delta, jiang2023avatarcraft, kawar2023imagic}. These optimization-based methods are costly and slow, motivating recent exploration of the feed-forward paradigm. Instruct-Pix2Pix~\cite{brooks2023instructpix2pix} tunes the diffusion model along with a language model for generalized editing tasks. Diffstyler~\cite{huang2024diffstyler} learn a content and style-specific denoiser for disentanglement. FreeStyle~\cite{he2024freestyle} modulates the U-Net feature for training-free stylization. Concurrent with this exploration is the stylized image generation~\cite{hertz2024style,wang2024instantstyle,wang2024instantstylep,gao2024styleshot,chen2023controlstyle}. While related, they focus on a different setting where the style is extracted from an image and the content is a prompt.  
In this work, we focus on the structure and appearance control in text-driven stylization scenarios.

\textbf{Quantitative Evaluation of Style Transfer.} Quantitatively evaluating style transfer is a long-standing problem. Initial approaches repurposed low-level metrics, including Gram Loss~\cite{gatys2016image}, LPIPS~\cite{zhang2018unreasonable}, and FID~\cite{heusel2017gans}. However, recent literature found that these metrics are fundamentally incapable of capturing the holistic and semantic qualities of human artistic perception~\cite{bo2018computational, so2023measuring, ioannou2024evaluation, chen2024learning, yi2023towards, wright2022artfid}. In response to these shortcomings, art-specific evaluators like ArtFID~\cite{wright2022artfid} and ArtScore~\cite{chen2024learning} were developed to better quantify the abstract concept of "artness". Nevertheless, they cannot handle open-vocabulary text-driven stylization and lack the mechanism to evaluate structure and appearance separately. This paper propose a semantic-level MLLM-based evaluator to assess the structure and appearance fidelity, which aligns better with human perception.
\section{Methodology}

\subsection{Objective: Disentangled Dual Controllability}

Given a source image $I$, a text-based style prompt $y$, and a pretrained diffusion model $\mathcal{G}(\cdot)$, our primary objective is to generate a stylized image $\hat{I}$ that preserves the semantic content of $I$ while harmoniously embodying the style described by $y$. The core innovation we pursue is \textbf{disentangled dual control}, meaning that decomposing the style prompt $y$ into two orthogonal components—structure and appearance—and controlling their strength independently. Our definition of structure and appearance in a 2D image is mainly based on fine art~\cite {gombrich1995story}. A formal definition of them is challenging as it relates to visual semiotics~\cite{van2000handbook,goodman1976languages}, extending beyond the scope of this paper. Generally speaking, structure corresponds to the shapes, like contours and curvatures, while appearance corresponds to local patterns, like strokes and color palettes. We also aim to develop an evaluator $\mathcal{E}$, which can evaluate the fidelity of structure and appearance style in a way aligned with human perception.

The remaining parts of this section are organized as follows. In Sec.~\ref{sec:prelim}, we review the basics of inversion-based image manipulation. In Sec.~\ref{sec:cs_noise_space},~\ref{sec:delegate_overview}, we explain the motivation and design of control at a high level, and details are described in Sec.~\ref{sec:cs_representation}. Sec.~\ref{sec:vlm_eval} outlines the proposed MLLM-based structure and appearance evaluators.

\subsection{Preliminary: DDIM Inversion}\label{sec:prelim}

 To stylize a source image $\mathbf{x}_0 := I$ using DMs, inversion-based methods first approximate the noise latents $\mathbf{x}_{1:T}$ of $I$, achieved via techniques such as DDIM inversion~\cite{song2020denoising}.  Stylization is then performed through re-generation with altered conditions (usually specified as a style prompt $y$). Specifically, one may start with an intermediate step $\mathbf{x}_\tau$ (i.e., control point), where $\tau \in [1, T]$ for iterative DDIM sampling. Each denoising step is formulated as follows:

\begin{align}
\mathbf{x}_{t-1} &= \sqrt{\alpha_{t-1}} \left( \frac{\mathbf{x}_t - \sqrt{1 - \alpha_t} \epsilon_\theta(\mathbf{x}_t, t; y)}{\sqrt{\alpha_t}} \right) \nonumber \\
&\qquad + \sqrt{1 - \alpha_{t-1}} \epsilon_\theta(\mathbf{x}_t, t; y),
\label{eqn:ddim_reverse}
\end{align}
where $\epsilon_\theta$ is the denoiser, $t$ is the timestep,  and $\alpha_{1:T}$ is a predefined noise schedule. The assumption of this paradigm is that with a proper $\tau$, the resulting stylized image $\hat{I} := \hat{\mathbf{x}}_0$ harmoniously integrates the structure and appearance of style in prompt $y$ with the source image $I$.

\subsection{Structure and Appearance in Noise Space}\label{sec:cs_noise_space}
Prevailing neural stylization methods are built on a paradigm that separates an image into ``content'' and ``style''~\cite{li2017demystifying,huang2017arbitrary,wang2023stylediffusion,kotovenko2019content,sanakoyeu2018style,zhang2023unified,zhang2020unified,ding2024domain}. In this view, the style usually refers to the feature maps extracted from certain layers of a neural network. To advance stylization towards both structure and appearance controllability, we adopt different modeling that decomposes an image as its structure and appearance components: $\mathbf{x}_0=\mathcal{G}_0(\mathbf{z}^s_0,\mathbf{z}^a_0)$, where $\mathcal{G}_0=(\cdot,\cdot)$ is a composition function, $\mathbf{z}^s_0$ and $\mathbf{z}^a_0$ are the latent structure and appearance factorization, respectively. This is a ``static'', image-level perspective. In the diffusion process, the distribution of image $x_0$ is tied with the intermediate distributions in $\mathbf{x}_{1:T}$, where the denoiser $\epsilon_\theta$ learns the transition $q_\theta(x_{t-1}|x_{t},y)$ via $\epsilon$-prediction. Therefore, we posit similar factorization of predicted noise residual, which is a ``dynamic'' decomposition across the full frequency bands: 

\begin{equation}
    \epsilon_\theta(\mathbf{x}_t, t; y) = \mathcal{G}_t\left(\kappa_t,\psi_t\right), \quad t\in[0,T]
    \label{eqn:entangle}
\end{equation}
where $\kappa_t$ and $\psi_t$ denote the structure and appearance representation at diffusion time-step $t$ (detailed later), respectively. $\mathcal{G}_t(\cdot,\cdot)$ is a conceptual noise-space composition function at time $t$.

\subsection{Structure and Appearance as Delegate Diffusion Process}\label{sec:delegate_overview}

We argue that the fundamental obstacle to achieving dual control in diffusion-based stylization is the inherent entanglement of structure and appearance. Our analysis, detailed in Appendix~\ref{append: analysis}, pinpoints the source of this problem: the reliance on a single latent trajectory $\mathbf{x}_\tau \xrightarrow{} \mathbf{x}_0$. This monolithic generation process forces structural and appearance attributes to compete for influence at every denoising step, creating the S-A tradeoff that fundamentally limits controllability.

To break this bottleneck, we propose a novel mechanism that stylizes an image with \textbf{separate} diffusion trajectories, as illustrated in Fig.~\ref{fig:overview}. Specifically, we leverage two supplementary diffusion processes with shared information, called \textit{delegate branches}. We initialize the structure and main branch from the inverted noise $\mathbf{x}_T$, while appearance delegation starts from a Gaussian $\epsilon \sim \mathcal{N}(\mathbf{0},\mathbf{I})$. These delegations enable controlling the stylization over the entire diffusion process. The controlled main branch can be denoted as:

\begin{equation}
\epsilon^m_\theta(\mathbf{x}_t, t; y, \kappa^s_t,\psi^a_t) = \mathcal{G}_t\left(\kappa^s_t\circ \kappa^m_t, \psi^a_t\diamond \psi^m_t \right),\quad t\in[0,T]
\label{eqn:controlled}
\end{equation}
where the superscripts $s$, $a$, and $m$ denote the factorization extracted from the structure, appearance delegation, and main branch, respectively. The $\circ$ and $\diamond$ are two non-commutative control operators.

\subsection{Structure and Appearance Representations in Denoising U-Net}\label{sec:cs_representation}

Having established the control mechanism in Eq.~\ref{eqn:controlled}, we now formulate the $\kappa$ and $\psi$ in a U-Net-based denoiser for disentangling structure and appearance control.

\begin{figure}
    \centering
    \includegraphics[width=0.8\linewidth]{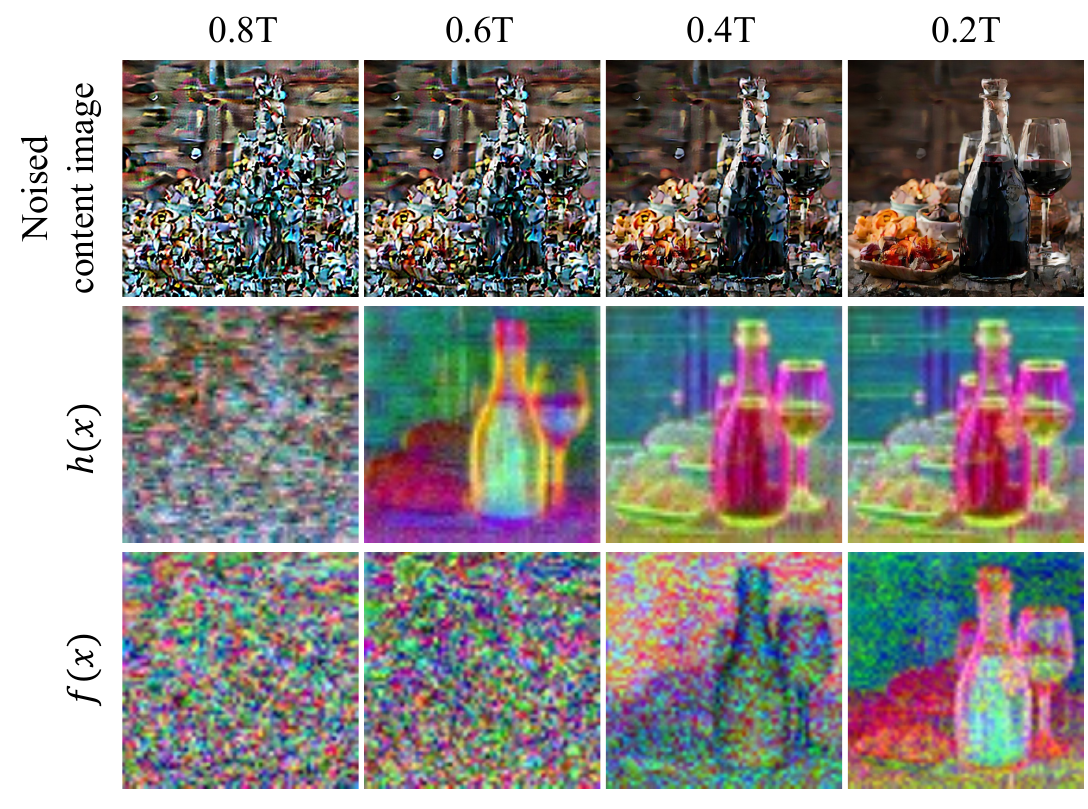}
    \caption{\textbf{ResBlock feature map visualization}. We apply t-sne to visualize the feature map of different feature maps in U-Net decoder. The hidden features $h(\mathbf{x})$ better preserves the semantics than the ResNet feature $f(\mathbf{x})$ throughout all $T$.}
    \label{fig:feat_vis}
\end{figure}

\textbf{Pyramidal Structure Representation $\kappa$.} To effectively control structural stylization, we require a representation that captures image semantics at multiple levels of abstraction. We identify the hidden features in the ResBlock of denoising U-Net as the ideal substrate for this purpose, which robustly encode appearance-invariant image semantics across varying $t$ (see Fig.~\ref{fig:feat_vis}). Formally, we denote the hidden feature of a ResBlock as $h_i(\mathbf{x}_t)$, where $i \in \{1, 2, \ldots, N_{res}\}$ indexes the ResBlocks up to $N_{res}$, with increasing spatial resolution. Stacking such feature from all layers forms a pyramidal structure representation of $\mathbf{x}_0$ at $t$:
\begin{equation}
\begin{aligned}
    \kappa^s_t &:= \{h_i(\mathbf{x_t})\}_{i\in S_{res}
}, \\
    \text{where } h_i &\text{ extracted from } \epsilon^s_\theta(\mathbf{x}_t, t; \varnothing), \\
\text{and } S_{res} &\subseteq \{1, 2, \ldots, N_{res}\}.
\end{aligned}
\end{equation}

Our representation is distinct as it captures multi-scale semantics and provides continuous guidance across the full denoising trajectory ($t \in [0,T]$). This fundamentally differs from methods relying on solitary control points (e.g., $\mathbf{x}_\tau$) or single-scale conditions (e.g., ControlNet, IP-Adapter). Such approaches are constrained to a fixed resolution and/or SNR, which architecturally limits their ability to generate complex structural styles. As evidenced in Sec.~\ref{sec: analysis}, this constraint often leads to undesirable semantic trade-offs.  With this pyramidal representation, we implement the structure control operator $\circ$ as \textit{injection} (i.e., $a\circ b = a$). 

\textbf{Semantic-aware Appearance Representation $\psi$.} We represent the appearance of the target style as self-attention maps extracted from all layers of $\epsilon^s_\theta$. For the style to be applied harmoniously, its generation must be guided by the image's semantics. However, until now, we denoise appearance delegation from Gaussian noise and hence has no information-sharing with the source image. To compensate for this, we propose \textbf{Structure-to-Appearance injection (S2A)} that propagates the high-level semantics into appearance generation. Specifically, we inject the self-attention value $V$ from early layers of $\epsilon^s_\theta$ to $\epsilon^a_\theta$. Let $N_{attn}$ denote the total number of attention blocks within the U-Net decoder, $S_{s2a} \subseteq \{1, 2, \ldots, N_{attn}\}$ be the selected blocks for S2A injection. The appearance representation at $t$ is:

\begin{equation}
\label{eq:s2a_definition}
\begin{aligned}
    \psi^a_t := \{A^a_i\}_{i=1}^{N_{attn}}, \quad \text{where } & A^a_i \text{ is extracted from } \epsilon^a_\theta \left( \mathbf{x}_t, t; y, \{V^s_j\}_{j\in S_{s2a}} \right), \\
    \text{with } & \{V^s_j\}_{j\in S_{s2a}} \text{ extracted from } \epsilon^s_\theta(\mathbf{x}_t, t; \varnothing).
\end{aligned}
\end{equation}

\begin{figure*}[!t]
    \centering
    \includegraphics[width=\linewidth]{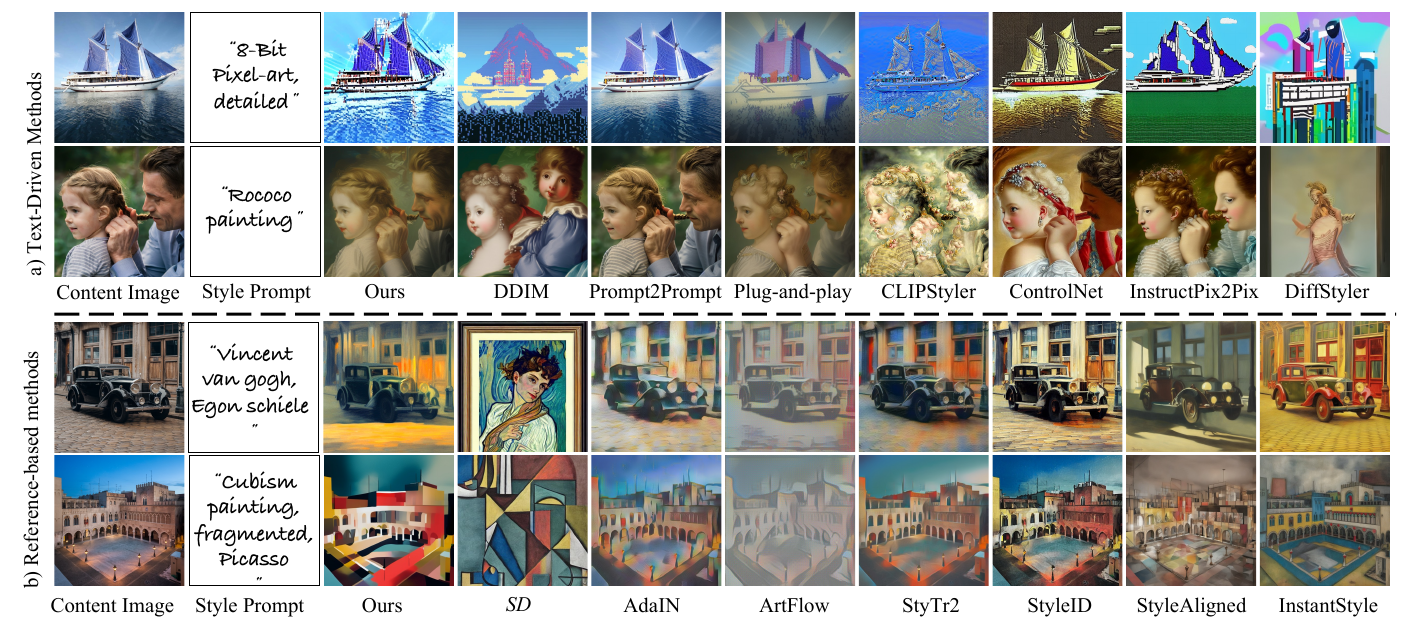}
    \caption{\textbf{Qualitative comparison with existing methods.} We compare our work with representative text-driven image manipulation method in (a), and image-based stylization methods in (b). Stylized images generated by DiffArtist produce high-fidelity structural and appearance-level style with semantic integrity. We suggest readers for more visualizations in Appendix~\ref{append: result}.}
    \label{fig:qual_cmp}
\end{figure*}

Inspired by StyleAligned~\cite{hertz2024style} we design the appearance-style control operator $\diamond$ as the AdaIN~\cite{huang2017arbitrary},
\begin{equation}
\label{eq:adain}
a \diamond b = \sigma(b)\left(\frac{a - \mu(a)}{\sigma(a)}\right) + \mu(b).
\end{equation}

We visualize the feature interactions in the right part of Fig.~\ref{fig:overview}. Adjusting the control layer in $S_{res}$ and style strength of $\epsilon_t^a$ enables disentangled control for structure and appearance, respectively.

\subsection{Structure and Appearance Evaluation via MLLMs}\label{sec:vlm_eval}
Recent research demonstrates the powerful semantic-level multimodal understanding of MLLMs~\cite{zhang2024mm,kojima2022large,zhang2023multimodal,li2025m2iv,li2025taco}. We leverage state-of-the-art MLLMs as a zero-shot evaluator to assess two key axes of our method: structure preservation and appearance fidelity. Crucially, our goal is to measure fidelity of stylized images, not subjective qualities like visual appeal.\footnote{Note that visual-appeal is also not equivalent to \textit{aesthetic quality}~\cite{gombrich1995story,jiang2025multimodal}}.

Given the inherent subjectivity and the difficulty of assigning absolute scores, we design a relative evaluation framework. Specifically, we query MLLM with the tuple $(\mathcal{\hat{I}}, I, y,y_{i})$, where $\mathcal{\hat{I}}=\{\hat{I}_1,\dots ,\hat{I}_k\}$ is the stylization result generated by $k$ different models, $I\coloneq\mathbf{x}_0$ is the source image, and $y_i$ is the instruction dedicated for structure or appearance fidelity evaluation. The MLLM is tasked with ranking the outputs of $k$ different methods for each criteria. We show in Sec.~\ref{sec:human_align} that this design achieves superior alignment with human perception compared with existing metrics.

\section{Experiments}
\begin{table*}[h]
\caption{\textbf{Quantitative comparison against existing methods.} We show conventional (in \textcolor{gray}{gray} font) and MLLM-based metrics for representative methods. For each metric, \fcolorbox{black}{best}{\rule{0pt}{4pt}\rule{4pt}{0pt}} indicates the best score, \fcolorbox{black}{second}{\rule{0pt}{4pt}\rule{4pt}{0pt}} indicates the second best score , and \fcolorbox{black}{third}{\rule{0pt}{4pt}\rule{4pt}{0pt}} indicates the third best score (best viewed in color). Win rate means the percentage that our method wins in pair-wise comparison.}

\centering
\resizebox{\textwidth}{!}{%
\begin{tabular}{lcccccccccc}
\toprule
\textbf{Metric} & \textbf{Ours} & \textbf{DDIM} & \textbf{SD} & \textbf{PnP} & \textbf{P2P} & \textbf{InstructP2P} & \textbf{ControlNet} & \textbf{InstantStyle} & \textbf{DiffStyler} & \textbf{CLIPStyler}\\
\midrule

\textbf{Inference time (sec)} & 10.5 & 9.7  & 3.9 & 55.3 & 29.1 & 9.2 & 7.8 & 7.8 & 18.2 & 24.2 \\

\textbf{Training~\&~adapter free} & \cmark{} & \cmark{} & \cmark{} & \cmark{} & \xmark{} & \xmark{} & \xmark{} & \xmark{} & \xmark{} & \xmark{} \\
\midrule

\textcolor{gray}{\textbf{LPIPS} $\downarrow$} & 
\textcolor{gray}{0.52} & \textcolor{gray}{0.57} & \textcolor{gray}{0.76} & \textcolor{gray}{0.67} &  \cellcolor{third}\textcolor{gray}{0.47} & \cellcolor{best}\textcolor{gray}{0.42} & \textcolor{gray}{0.65} & \textcolor{gray}{0.59} & \textcolor{gray}{0.71} & \cellcolor{second}\textcolor{gray}{0.46} \\

\textcolor{gray}{\textbf{CLIP Score}~\cite{radford2021learning} $\uparrow$} &
\cellcolor{third}\textcolor{gray}{25.91} & \textcolor{gray}{25.25} & \cellcolor{best}\textcolor{gray}{27.46} & \textcolor{gray}{24.89} & \textcolor{gray}{23.48} & \textcolor{gray}{21.94} & \textcolor{gray}{24.93} & \textcolor{gray}{22.85} & \textcolor{gray}{25.79} & \cellcolor{second}\textcolor{gray}{27.14} \\

\textcolor{gray}{\textbf{PickScore}~\cite{kirstain2023pick} $\uparrow$} &
\cellcolor{third}\textcolor{gray}{20.51} & \cellcolor{second}\textcolor{gray}{20.58} & \cellcolor{best}\textcolor{gray}{20.68} & \textcolor{gray}{20.34} & \textcolor{gray}{20.50} & \textcolor{gray}{20.06} & \textcolor{gray}{20.46} & \textcolor{gray}{19.97} & \textcolor{gray}{19.24} & \textcolor{gray}{20.13} \\
\midrule
\textbf{Structure (MLLM) $\uparrow$} & 
\cellcolor{second}0.61 & 0.22 & 0.29 & 0.52 & \cellcolor{best}0.65 & \cellcolor{third}0.60 & 0.58 & 0.56 & 0.35 & 0.51 \\

\textbf{Appearance (MLLM) $\uparrow$} & 
\cellcolor{best}0.67 & 0.46 & 0.31 & \cellcolor{second}0.60 & 0.47 &\cellcolor{third}0.59 & 0.55 & \cellcolor{best}0.67 & 0.30 & \cellcolor{third}0.59 \\

\textbf{Avg. (MLLM) $\uparrow$} & 
\cellcolor{best}0.64 & 0.34 & 0.30 & 0.56 & 0.56 & 0.60 & \cellcolor{third}0.57 & \cellcolor{second}0.62 & 0.33 & 0.55 \\

\midrule
\textbf{Structure Win (Human) $\uparrow$} & - & 78.2\% & 62.4\% & 64.7\% & 57.3\% & 62.2\% & 71.2\% & 59.8\% & 81.3\% & 73.0\% \\
\textbf{Appearance Win (Human) $\uparrow$} & - & 74.2\% & 86.4\% & 62.0\% & 73.7\% & 68.7\% & 75.0\% & 60.1\% & 85.3\% & 76.3\% \\

\bottomrule
\end{tabular}%
}

\label{tab:comp}
\end{table*}

\subsection{Experiment Setup} 

\textbf{Implementation Details.} Our experiments are built upon the publicly available Stable Diffusion 2.1 model\footnote{\url{https://huggingface.co/stabilityai/stable-diffusion-2-1}}. We perform DDIM sampling with \(T = 50\) steps. During the inversion, we record the intermediate noise predictions to overwrite the input of $\epsilon^s_\theta$ during denoising. Further implementation details are available in Appendix~\ref{append: detail}. Experiments were conducted with a single RTX 4090-D GPU, with an approximate runtime of 2 seconds for inversion and 8 seconds for the final stylization.

\textbf{Default Parameters.} In main experiment, we default \(S_{\text{res}}\) to be the first four ResNet layers (\verb|[1, 2, 3, 4]|), and \(S_{\text{s2a}}\) as the first two attention layer features (\verb|[1, 2]|). The classifier-free guidance (CFG) scale is set as $7.5$. These default control parameters correspond to moderate structural and appearance variations, used to set a fair comparison with existing works to avoid per-image parameter tuning. However, it should be noted that users can adjust these parameters for customization.

\textbf{Compared Methods.}
We compare our method against existing text-driven stylization and manipulation methods: DDIM Inversion~\cite{song2020denoising}, CLIPStyler (optimization-based)~\cite{kwon2022clipstyler}, DiffStyler~\cite{huang2024diffstyler}, Plug-and-Play (PnP), Prompt2Prompt (with null text inversion) (P2P) ~\cite{mokady2023null,hertz2022prompt}, ControlNet~\cite{zhang2023adding}, and InstructPix2Pix~\cite{brooks2023instructpix2pix}. Additionally, we consider a baseline named \textit{SD}, which generates images with Stable Diffusion according to $y$. 

We also indirectly compare our method with reference-based stylization methods, including AdaIN~\cite{huang2017arbitrary}, ArtFlow~\cite{an2021artflow}, StyTr2~\cite{deng2022stytr2}, StyleID~\cite{chung2024style}, StyleAligned~\cite{hertz2024style} (with ControlNet), and InstantStyle~\cite{wang2024instantstyle} (with ControlNet). Images generated by \textit{SD} are used as reference.

\textbf{Conventional Metrics}: \textbf{LPIPS}~\cite{zhang2018unreasonable} measures the content preservation by calculating the feature distance between the source and stylized image. For style fidelity, we leverage \textbf{CLIP Score}~\cite{radford2021learning} and \textbf{Pick Score}~\cite{kirstain2023pick}, both of which quantify the alignment between the stylized image $\hat{I}$ and prompt $y$. We also include a \textbf{human study} crowd-sourced from $n_1=200$ users and report the average preference rate for our method.

\textbf{MLLM-based Metrics:} We prompt the MLLM to rank the fidelity of $k$ stylized images from best (rank 1) to worst (rank $k$). We normalize the integer ranking and average it over the whole evaluation set. Therefore, a score closer to $1$ indicates a stronger fidelity. We use \verb|Gemini-v2.0-flash| for its strong multimodal capability. The full prompt templates can be found in Appendix~\ref{append: MLLM}.

\subsection{Comparisons}

\textbf{Qualitative Comparisons.} We first provide a comprehensive comparison against previous methods, visualized in Fig.~\ref{fig:qual_cmp}-(a, b).  \textbf{(a)}: Compared with text-driven methods, DiffArtist is the best at following the style prompt while maintaining semantic integrity. Our method enables harmonious structural variations, such as pixelation, without compromising intricate details like facial identity and hair.  By contrast, the compared methods may produce misaligned styles (\textit{e.g.,} CLIPStyler, Plug-and-Play) or introduce undesired modifications that violate semantics (\textit{e.g.,} DiffStyler, ControlNet). \textbf{(b)}: When broadly compared with reference-based methods, DiffArtist still stands out for its high stylization fidelity from two perspectives. To fully demonstrate the superiority, we \textbf{highly suggest readers for additional visualizations} in Appendix~\ref{append: result}.

\textbf{Quantitative Comparison.} For our quantitative evaluation, we first sample 50 art styles from WikiART, with both abstract (e.g., ``Cubism'') and realistic styles (e.g., ``High Renaissance''), which are further diversified by GPT-4o in terms of description. This diversification sets a broad spectrum of styles to align with real-world user inputs. The content comprises 50 images from MSCOCO~\cite{lin2014microsoft} and 50 photorealistic images generated by another model~\cite{ideogram2024}. For each of the $100$ content images, we randomly draw $10$ style prompts from all possible styles, resulting in a total of $1,000$ unique combinations for comparison.  Tab.~\ref{tab:comp} presents the results.

For conventional metrics, DiffArtist achieves an LPIPS of 0.52, a CLIP Score of 25.91, and a PickScore of 20.51, outperforming most of the compared methods. However, these metrics do not measure stylization quality in structure and appearance. As a simple counter-example, the baseline \textit{SD} has the highest CLIP Score and PickScore, whereas it is not even performing stylization. We include these scores solely for reference.

When evaluated with MLLM-based metrics, DiffArtist attains the highest average score of 0.64. Specifically, our method achieves the second-highest structure score of 0.61, demonstrating DiffArtist's effectiveness in generating structural styles. While the editing-focused P2P method scores higher, it does so by sacrificing stylistic strength, evidenced in qualitative comparisons. Besides, our method achieves the best appearance fidelity score, confirming its superior ability to render the appearance details from the text prompt. In human evaluations, our method is preferred by an average of 67.8\% of users in pairwise comparison, further validating its superiority.

\section{Analysis and Discussion}\label{sec: analysis}

This section provide in-depth analysis on the proposed system. In Sec.~\ref{sec:fine_compare}, we analyze the controllability of DiffArtist in detail. Sec.~\ref{sec:human_align} validates the effectiveness of MLLMs as style evaluators. Sec.~\ref{sec:ablate} provides ablations on delegations and the S2A injection. We conclude  with a discussion in Sec.~\ref{sec:limit}.

\begin{figure*}[!t]
    \centering
    \includegraphics[width=\linewidth]{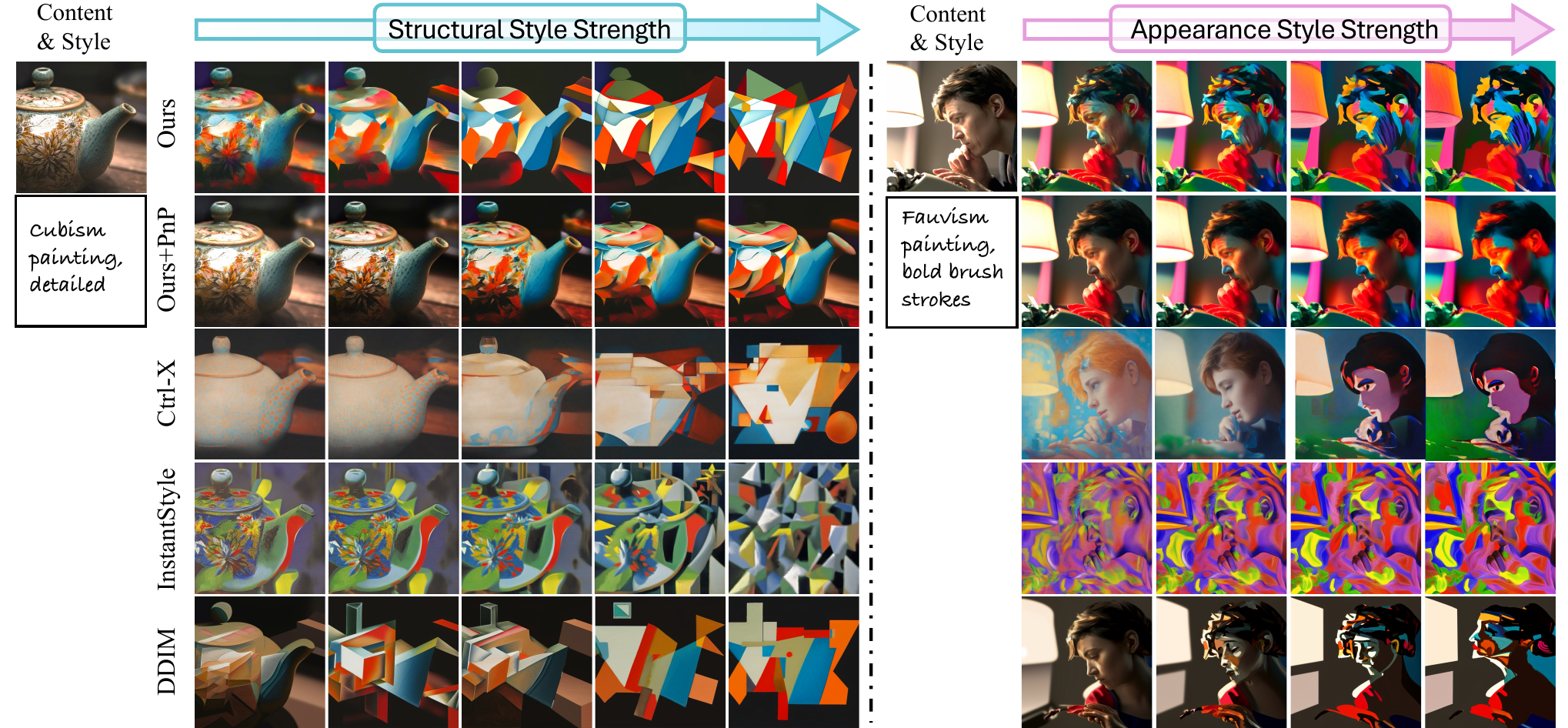}
    \caption{\textbf{DiffArtist offers disentangled and stronger controllability.}  (Left): DiffArtist enables smooth and artistically meaningful structural stylization at varying degree, without violating appearance style strength. (Right): DiffArtist allows fine-grained control of appearance-related style strength while preserving structural and semantic integrity. }
    \label{fig:control_cmp}
\end{figure*}

\subsection{DiffArtist has Strong Controllability}\label{sec:fine_compare}

This subsection validates the unprecedented dual controllability of DiffArtist. To achieve this, we conduct a fine-grained comparative analysis against representative methods, categorizing them by their core control mechanism: (a) \textbf{Semantic pyramid:} include DiffArtist, DiffArtist implemented with Plug-and-Play structure representation, $f(\mathbf{x})$ (Ours + PnP), and Ctrl-X; (b) \textbf{Pixel-level map:} include InstantStyle~\cite{cui2023instastyle}, which is based on ControlNet~\cite{zhang2023adding}; (c) \textbf{Noise inversion}, which corresponds to the DDIM baseline. For structure control, we define five levels from weakest to strongest. For group (a) we use the following control layers: ($\varnothing$, \verb|[1],[1-4],[1-6],[1-8]|); for group (b), we evenly sample their respective control strength parameters; and we use $\tau=[0,5,10,15,20]$ for group (c). Appearance strength is controlled by sampling CFG weights in \verb|[2.5,5,7.5,10]| for all groups except for Ctrl-X, which is achieved by adjusting its appearance schedule parameter.

\textbf{Qualitative Comparison on Control.} As visualized in Fig.~\ref{fig:control_cmp}, DiffArtist demonstrates superior controllability, with harmonious, consistent, and disentangled interpolations across a sequence of control levels. It correctly captures the essential geometric principles of Cubism, whereas competing methods merely apply a superficial texture (e.g., Ours+PnP) or fail to produce meaningful structural variations (e.g., InstantStyle). This nuanced control is further evident when adjusting style strength; DiffArtist provides an artistically meaningful interpretation by producing bolder strokes according to style prompt, while other methods resort to simplistic and often undesirable increases in color saturation. Most critically, our approach preserves semantic integrity. It avoids the catastrophic failures of pixel-map methods like InstantStyle, which can render the face unrecognizable, and also prevents the facial structure corruption seen in inversion-based methods that inherently entangle form with appearance. The extended visualizations in Appendix~A  with diverse style and content further confirm the superiority of DiffArtist.

\textbf{Quantitative Comparisons on Control.} We provide quantitative experiments to substantiate our visual observations, evaluating control based on two properties: \textbf{fidelity} and \textbf{editability}. We measure the fidelity as structure and appearance MLLM scores at different control levels. The results are reported in Tab.~\ref{tab:ctrl_qualtity}. DiffArtist outperforms others significantly and consistently, demonstrating \textbf{superior control fidelity}.

The editability defines the quality of the manipulation itself, which we assess via three criteria: \textbf{range, monotonicity}, and \textbf{disentanglement}. An ideal control of stylization should cover a wide range of stylistic effects (a large spread in MLLM scores), exhibit predictable monotonicity (Spearman's $\rho \approx 1$ for the target attribute), and maintain disentanglement from other attributes. We measure the disentanglement using Kendall's $W$ on the off-target scores, where a stable, unaffected score sequence yields $W \approx 0$. For instance, when controlling structural strength, $W$ for the appearance score is desired to be near 0. As shown in Tab.~\ref{tab:ctrl_qualtity}, the control of DiffArtist is the most editable: it covers the broadest range of effects, demonstrates the strongest monotonicity, and achieves the best disentanglement, reaffirming the superior control visualized in Fig.~\ref{fig:control_cmp}.

\begin{table}

\centering
\caption{Fidelity of structure and appearance control via cross-method comparison. The values correspond to the structure or appearance score (MLLM, $\uparrow$).  Note that the magnitudes of scores are only comparable within each column. The best result for each column is in \textbf{bold}.}
\resizebox{\columnwidth}{!}{
\begin{tabular}{cccccccccc}
\toprule
\multirow{2}{*}{Method} & \multicolumn{5}{c}{$\text{\textemdash}$~\textcolor{s_color}{Structure} $\rightarrow$} & \multicolumn{4}{c}{$\text{\textemdash}$~\textcolor{a_color}{Appearance} $\rightarrow$} \\
\cmidrule(lr){2-6} \cmidrule(lr){7-10}
 & lv.1 & lv.2 & lv.3 & lv.4 & lv.5 & lv.1 & lv.2 & lv.3 & lv.4 \\
\midrule
Ours & \textbf{0.62} & \textbf{0.65}&\textbf{0.74}& \textbf{0.63}&\textbf{0.66} & \textbf{0.70} & \textbf{0.74} &\textbf{0.80} & \textbf{0.78}\\
Ours+PnP & 0.43& 0.38 & 0.28 & 0.42&0.34 & 0.21 & 0.26& 0.27 & 0.32\\
Ctrl-X & 0.49&0.46 & 0.36& 0.46 &0.47 &0.41 & 0.37&0.33 & 0.42 \\
InstantStyle & 0.42& 0.42& 0.33 & 0.43 &0.35 & 0.34& 0.34&0.35 & 0.60 \\
DDIM & 0.49&    0.52& 0.52& 0.51& 0.55& 0.59& 0.53& 0.50& 0.52\\
\bottomrule
\end{tabular}
}
\label{tab:ctrl_qualtity}
\end{table}

\begin{table*}
\centering
\caption{Control editability and disentanglement via inter-method comparison. Higher $\rho$ indicates stronger monotonicity while lower $W$ means ranks are indistinguishable. When controlling from one perspective, a high $\rho$ is desired for editability, and a low $W$ for the other aspect is expected for disentangled control. The controls in DiffArtist are the most editable and disentangled. }
\resizebox{0.9\linewidth}{!}{
\begin{tabular}{ccccccccccccc}
\toprule
& \multicolumn{5}{c}{Sequential \textcolor{s_color}{Structure}-Control Only} & & \multicolumn{5}{c}{Sequential \textcolor{a_color}{Appearance}-Control Only} \\
\cmidrule(lr){2-6} \cmidrule(lr){8-12}
& Ours & Ours+PnP & Ctrl-X & InstantStyle & DDIM & & Ours & Ours+PnP & Ctrl-X & InstantStyle & DDIM \\
\midrule
$\rho(S) \uparrow$ & \textbf{0.82} & 0.54 & 0.32 & 0.39 & 0.70 & $W(S) \downarrow$ & 0.37  & \textbf{0.32}  & 0.36 & 0.45 &  0.69 \\
$W(A) \downarrow$& \textbf{0.32 } &  0.44& 0.45 & 0.34 & 0.72 & $\rho(A) \uparrow$ & \textbf{0.71} & 0.42 &  0.35 & 0.26 & 0.68 \\
\bottomrule
\end{tabular}
}
\label{tab:my_table}
\end{table*}

\subsection{MLLMs are Human-Aligned Stylization Evaluators}\label{sec:human_align}
We evaluate how each stylization metric aligns with human feedback by calculating the statistical correlation with the rankings from human feedback. To achieve this, we first construct a comparison set of 800 stylized images\footnote{The images evaluated here do not overlap with the main experiment in Tab.~\ref{tab:comp}}, and compare how human and MLLM preferences correlate.

To gather human feedback, we consider two groups of users. For the non-expert group, we recruited a large-scale group of $n_1 = 200$ participants through a crowdsourcing platform. Each participant performed a series of randomly sampled ranking tasks. To ensure the integrity of the collected data, we implemented attention checks and consistency filters to remove unreliable responses. We also recruited an expert group of $n_2 = 12$ participants with knowledge of fine art.

We measured the alignment between each metric's rankings and the human-derived preferences using Spearman's rank correlation ($\rho$). The averaged (of all content-style pairs) $\rho$ for both groups is reported in Tab.~\ref{tab:vlm_metric}. As the table shows, the MLLM-based metrics are \textbf{better aligned with human perception}, validating its effectiveness as an evaluation metric for style transfer.

\begin{table}[]
\centering
\caption{\textbf{Metrics correlation with human feedback.} We report correlation $\rho$ and combined significance $p$. The MLLM scores show stronger alignment with both expert and non-expert users.}
\begin{adjustbox}{width=\linewidth}
\begin{tabular}{cl|cc|cc}
\toprule
& & \multicolumn{2}{c|}{\textbf{Corr.~(Non-expert)}} & \multicolumn{2}{c}{\textbf{Corr.~(Expert)}} \\
\cmidrule(lr){3-4} \cmidrule(lr){5-6}
& \textbf{Metrics} & $\rho$ $\uparrow$ & p-value $\downarrow$ & $\rho$ $\uparrow$ & p-value $\downarrow$ \\
\midrule
\multirow{3}{*}{\textbf{\textcolor{s_color}{S}}} 
    & SSIM~\cite{wang2004image} & 0.29 & 0.12 & 0.25 & 0.14  \\
    & MLLM (GPT-4o) & 0.44 & 0.004 & 0.34 & 0.20  \\
    & MLLM (Gemini 2.0) & \textbf{0.42} & 0.02 & \textbf{0.45} & 0.03  \\
\midrule
\multirow{4}{*}{\textbf{\textcolor{a_color}{A}}} 
    & CLIP Score & 0.05 & 0.73 & 0.01 & 0.75 \\   
    & Pick Score~\cite{kirstain2023pick} & 0.27 & 0.11 & 0.25 & 0.13 \\   
    & MLLM (GPT-4o) & 0.25 & 0.05 & 0.22 & 0.06  \\
    & MLLM (Gemini 2.0) & \textbf{0.48} & 0.04 & \textbf{0.41} & 0.02  \\
\bottomrule
\end{tabular}
\end{adjustbox}

\label{tab:vlm_metric}
\end{table}

\subsection{Ablations}\label{sec:ablate}
\textbf{The delegations enable dual controllability.} DiffArtist's controllability stems from delegating structure and appearance generation to separate processes. To test the necessity of each, we created two ablated variants for comparison where the structure or appearance delegation is removed. Tab.~\ref{tab:ablate_c2s} presents the result of this ablation. The full methods achieves the best results, demonstrating the synergistic effect of delegations for dual controllability.

\begin{table}[t]
\centering
\caption{\textbf{Ablation on delegation branches.}  The proposed two delegations are complementary to each other, and the full methods achieves the highest fidelity.}
\begin{adjustbox}{width=\columnwidth}
\begin{tabular}{l|ccc}
\toprule
    {\diagbox{\textbf{Metric}}{\textbf{Method}}} & \textbf{full} & \textbf{w/o} \textcolor{s_color}{\textbf{structure}} & \textbf{w/o} \textcolor{a_color}{\textbf{appearance}}\\
    \midrule

    LPIPS $\downarrow$ & 0.51 & 0.76 & \textbf{0.42} \\
    CLIP Score $\uparrow$ & 25.91 & \textbf{27.69} & 21.75 \\
    Pick Score~\cite{kirstain2023pick} $\uparrow$ & 20.55 & \textbf{20.57} & 20.41 \\
    \midrule
    Structure (MLLM) $\uparrow$ & \textbf{0.72} & 0.37 & 0.33 \\
    Appearance (MLLM) $\uparrow$ & \textbf{0.62} & 0.59 & 0.22 \\
\bottomrule
\end{tabular}

\end{adjustbox}

\label{tab:ablate_c2s}
\end{table}

\begin{figure}
    \centering
    \includegraphics[width=\linewidth]{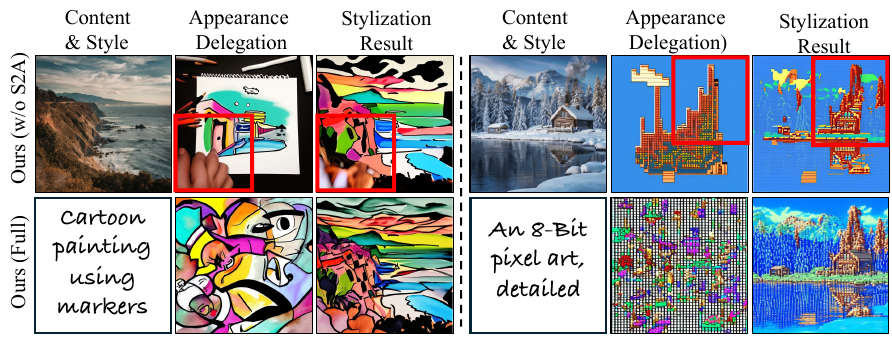}
    \caption{\textbf{Ablation on S2A injection.} S2A injection propagates the high-level semantic to the appearance generation. It avoids spatial misalignment of appearance-style strength.}
    \label{fig:c2s}
\end{figure}

\textbf{S2A injection} promotes semantic-aligned spatial distribution of style strength in the style delegation process, thereby avoiding artifacts in the final stylization result. We visualize the denoised style image (from appearance delegation) and the final stylization result in Fig.~\ref{fig:c2s}. Without S2A injection, the appearance delegation fails to align with content semantics, generating an appearance reference image with undesired patterns and an uneven texture distribution. These flaws manifest directly in the final output as distracting visual artifacts. In contrast, the full model leverages S2A to produce a coherent style representation, resulting in a clean and high-quality final image.

\subsection{Limitations \& Future work}
\label{sec:limit}

While DiffArtist marks a significant step towards disentangled structure and appearance control, we identify several limitations that open exciting avenues for future research. For instance, the structure control in DiffArtist is at a global level, and it cannot control the structure for each object separately. Many art styles exhibit mixed structure variation, like Surrealism and Collage art. Developing dense structure evaluators with 2D feedback signals is a promising direction~\cite{liang2024rich}, which may be further utilized as a reward model for reinforcement learning~\cite{black2023training,clark2023directly}.

\section{Conclusion}

We present the first exploration of structure- and appearance-controllable image stylization. Our contributions include DiffArtist, a styler that fully disentangles structure and appearance during the diffusion process, and a human-aligned evaluator to assess structural and appearance fidelity at the semantic level. Our extensive analysis proves that semantically-rich representations are essential for both structure and appearance style. We demonstrated that our design allows for high style fidelity and controllability, similar to that of a human artist. We believe the objective established in this paper—to stylize in both structure and appearance—offer a roadmap for the next generation of generative art tools to produce artistically meaningful paintings.
\section{Acknowledgment}
This research was supported by the Hong Kong Research Grants Council (GRF-15229423).

\bibliographystyle{ACM-Reference-Format}
\balance
\bibliography{main}

\clearpage
\appendix

\newpage
\begin{appendices}
\section{Additional Qualitative Results}
\label{append: result}
\begin{figure}[!t]
    \centering
    \includegraphics[width=\linewidth]{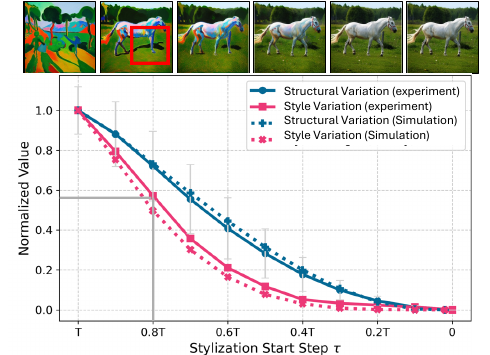}
\caption{\textbf{Tradeoff between Structure and Appearance Style Control.} We present the experimental (solid lines) and simulation (dotted lines) trends of structural and appearance variation in the diffusion process. Due to quadratic growth, highly noisy steps are required to achieve strong appearance styles, which are associated with significant structural variation, which can violate semantics. Top: Example stylization results starting from different steps, using the prompt: \textit{``Fauvism painting''}. When the appearance strength is high ($t=0.8T$), the structure (legs of the horse) is incorrectly modified.
}
    \label{fig:trajectory}
\end{figure}

\begin{figure}[t]
    \centering
    \includegraphics[width=\linewidth]{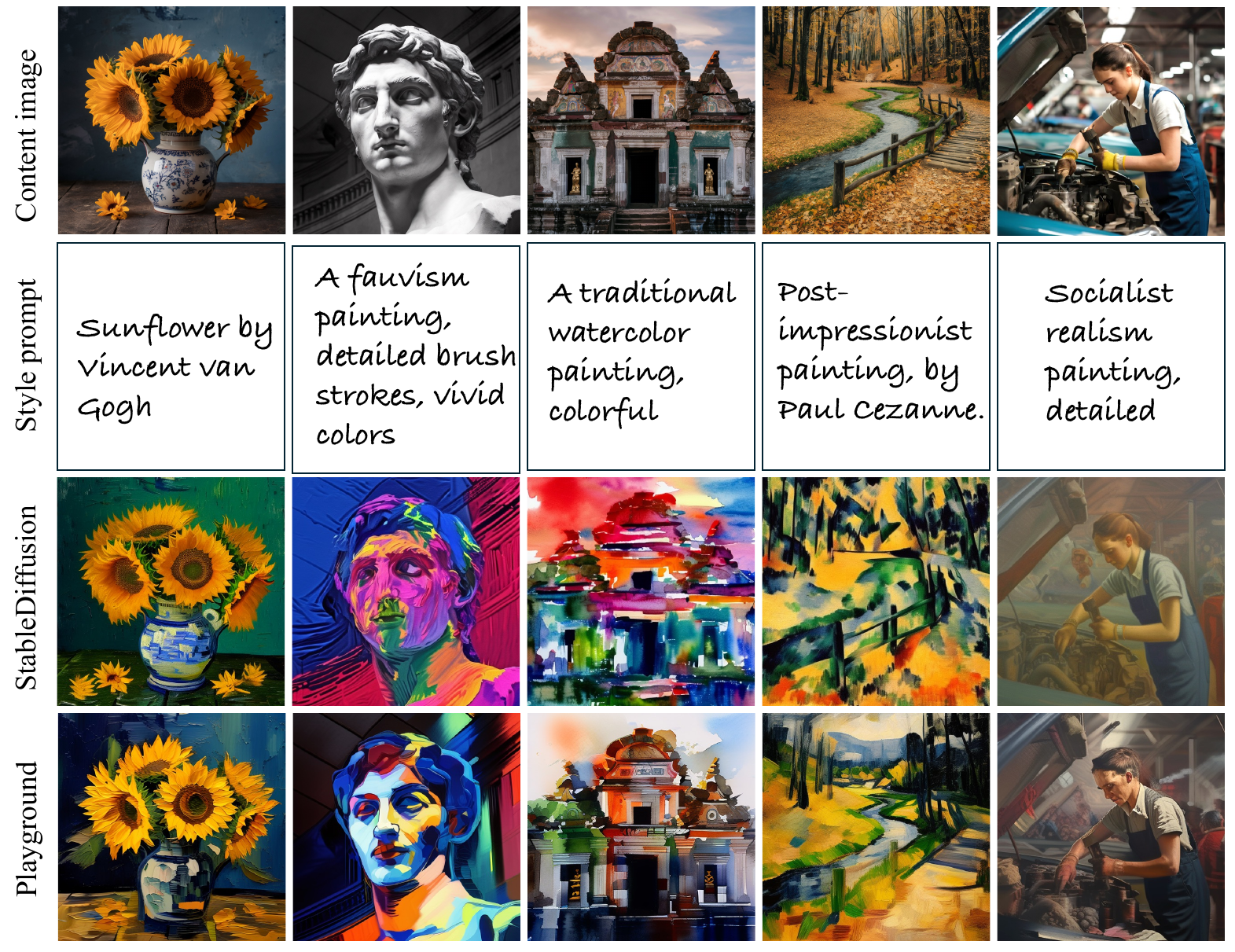}
    \caption{\textbf{DiffArtist implemented with different diffusion architecture.} We implement DiffArtist on the playground-v2 diffusion model. Similar stylization results could be achieved, demonstrating the generalizability of proposed method.}
    \label{fig:playground}
\end{figure}

\begin{figure}[]
    \centering
    \includegraphics[width=\linewidth]{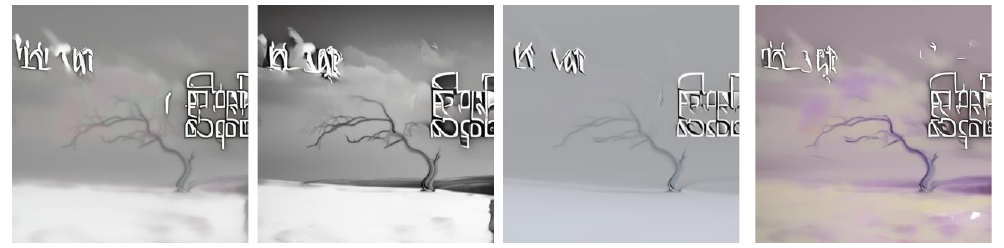}
    \caption{Example of failure cases.}
    \label{fig:fail}
\end{figure}

\begin{figure*}[!htbp]
    \centering
    \includegraphics[width=0.9\linewidth]{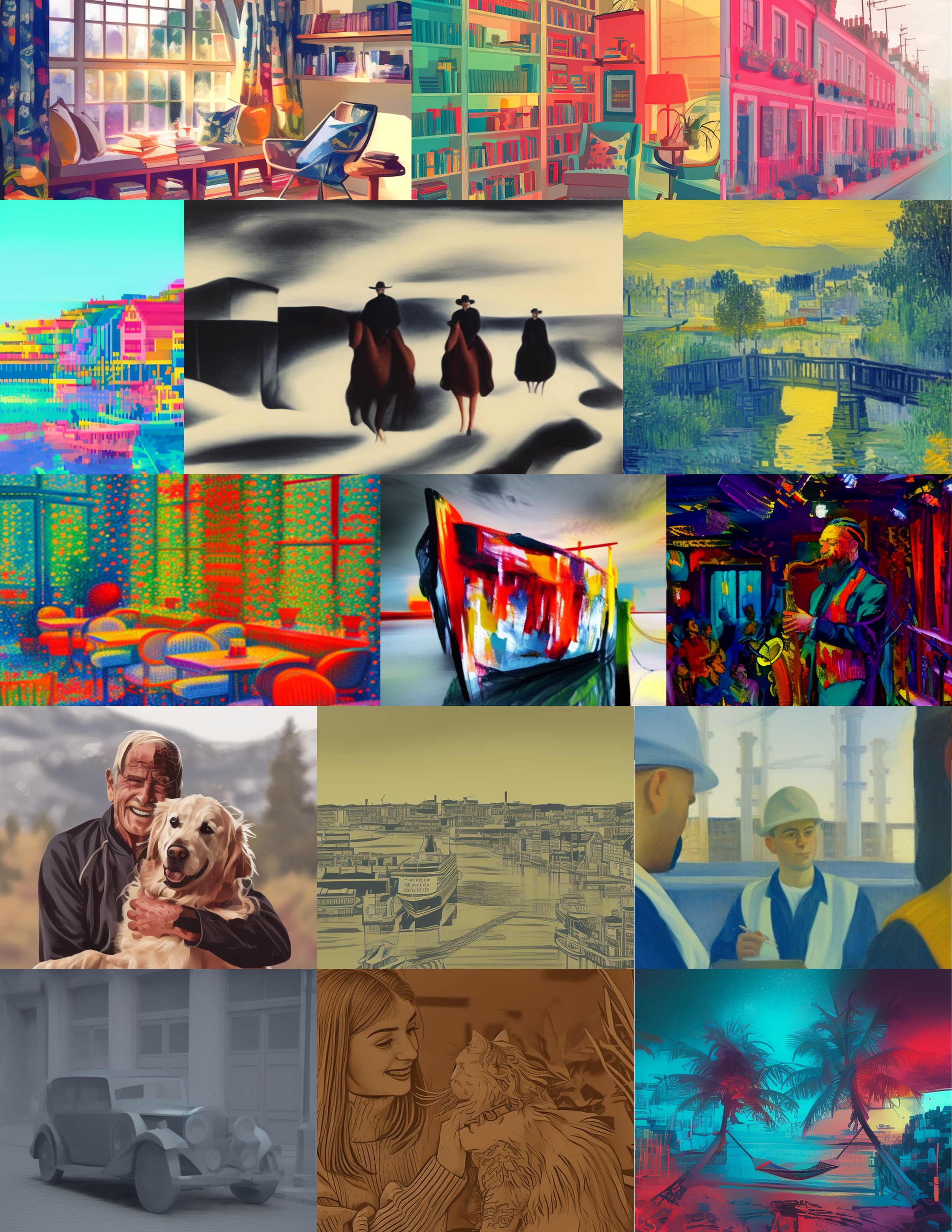}
    \caption{Addition results for DiffArtist (with default control parameters). The image semantics are preserved with strong and high-fidelity styles harmoniously integrated.}
    \label{fig:more_style_grid}
\end{figure*}

\begin{figure*}[!htbp]
    \centering
    \includegraphics[width=0.9\linewidth]{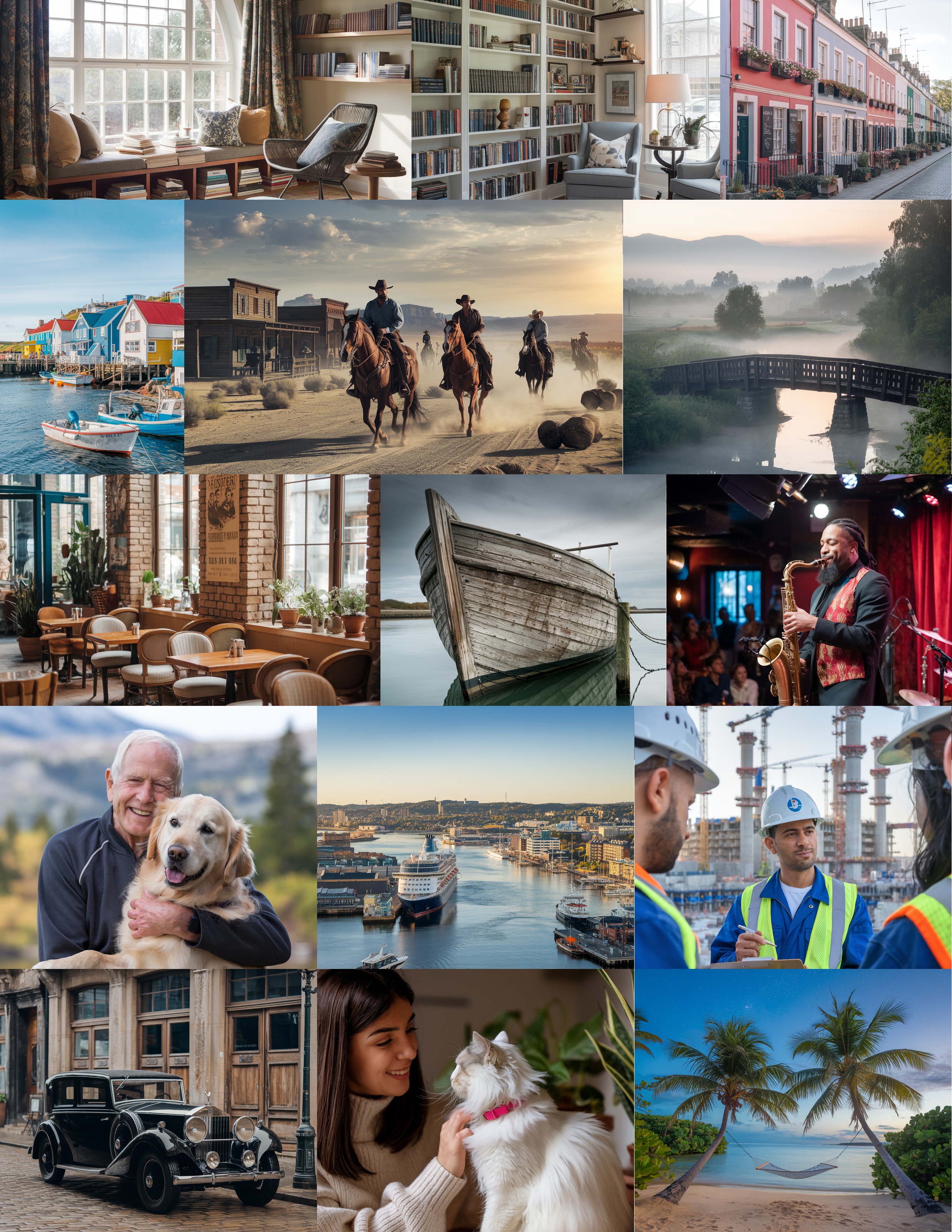}
    \caption{The corresponding source images of Figure ~\ref{fig:more_style_grid}.}
    \label{fig:more_style_grid_ref}
\end{figure*}

\begin{figure*}[!htbp]
    \centering
    \includegraphics[width=\linewidth]{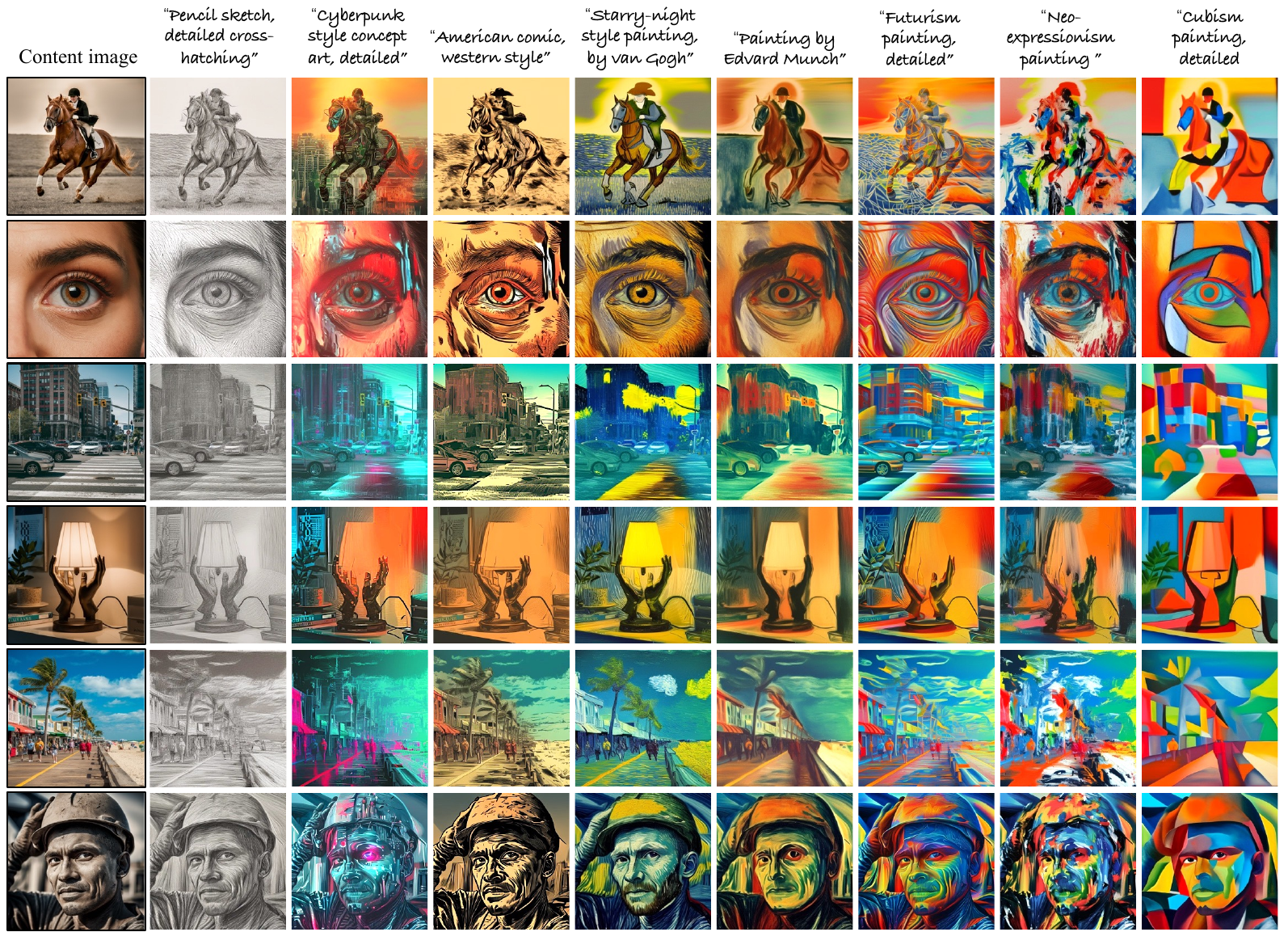}
    \caption{A grid experiment with different content and styles.}
    \label{fig:grid_teaser}
\end{figure*}

\begin{figure*}[!htbp]
    \centering
    \includegraphics[width=\linewidth]{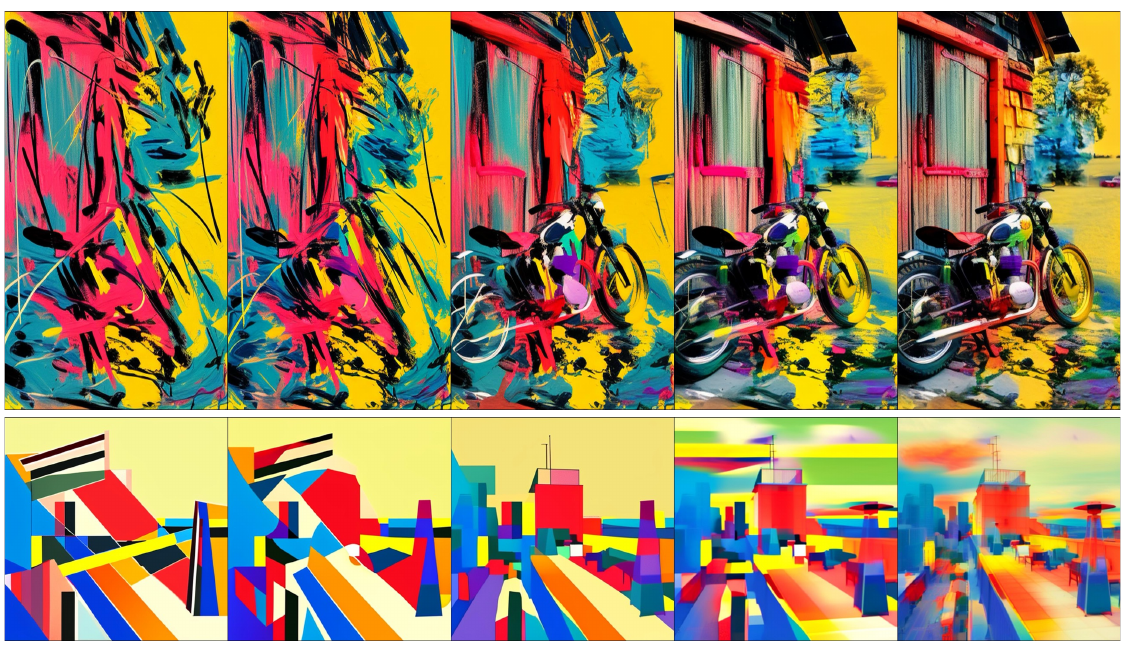}
    \caption{Additional result on structure control - 1.}
    \label{fig:more_structure_1}
\end{figure*}

\begin{figure*}[!htbp]
    \centering
    \includegraphics[width=\linewidth]{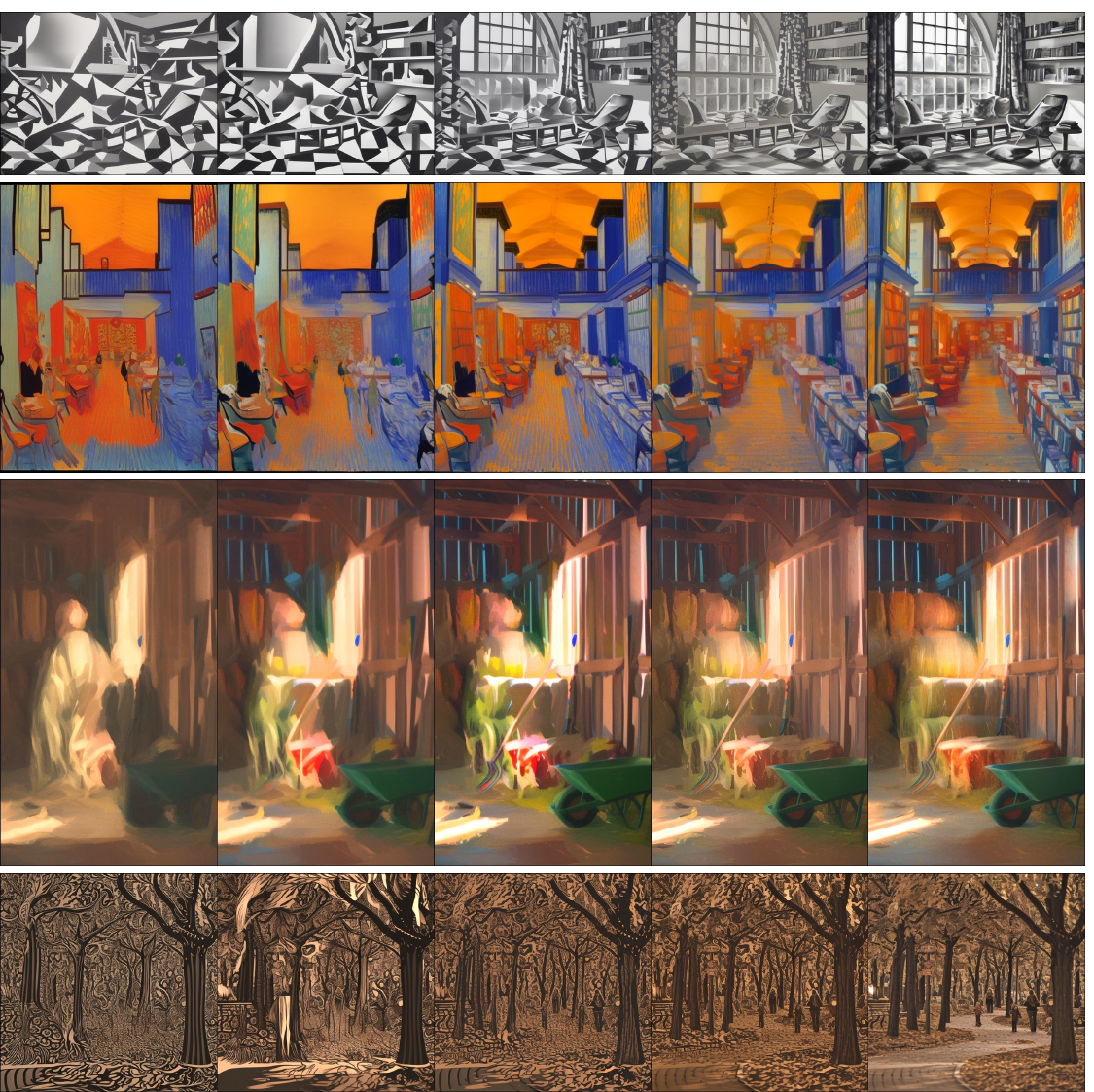}
    \caption{Additional result on structure control - 2.}
    \label{fig:more_structure_2}
\end{figure*}

\textbf{Visualization}
Additional appearance stylization is in Fig.~\ref{fig:more_style_grid}, the source image is in Fig.~\ref{fig:more_style_grid_ref}. A grid of different images with different styles is in Fig.~\ref{fig:grid_teaser}. Additional structure control is in Fig.~\ref{fig:more_structure_1} and Fig.~\ref{fig:more_structure_2}.

\begin{figure*}[!htbp]
    \centering
    \includegraphics[width=\linewidth]{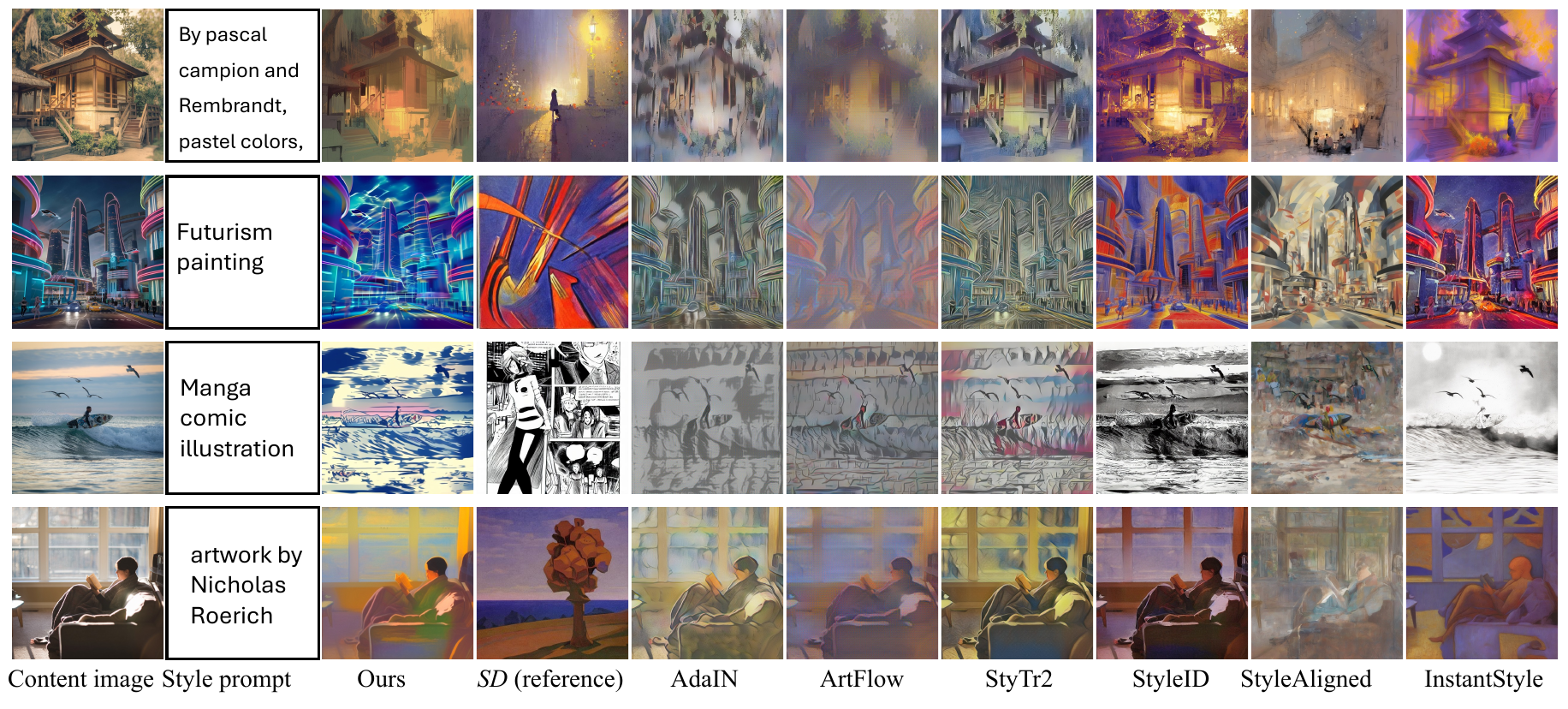}
    \caption{Extended comparison with reference-based style transfer methods.}
    \label{fig:supp_ref_cmp_1}
\end{figure*}

\begin{figure*}[!htbp]
    \centering
    \includegraphics[width=\linewidth]{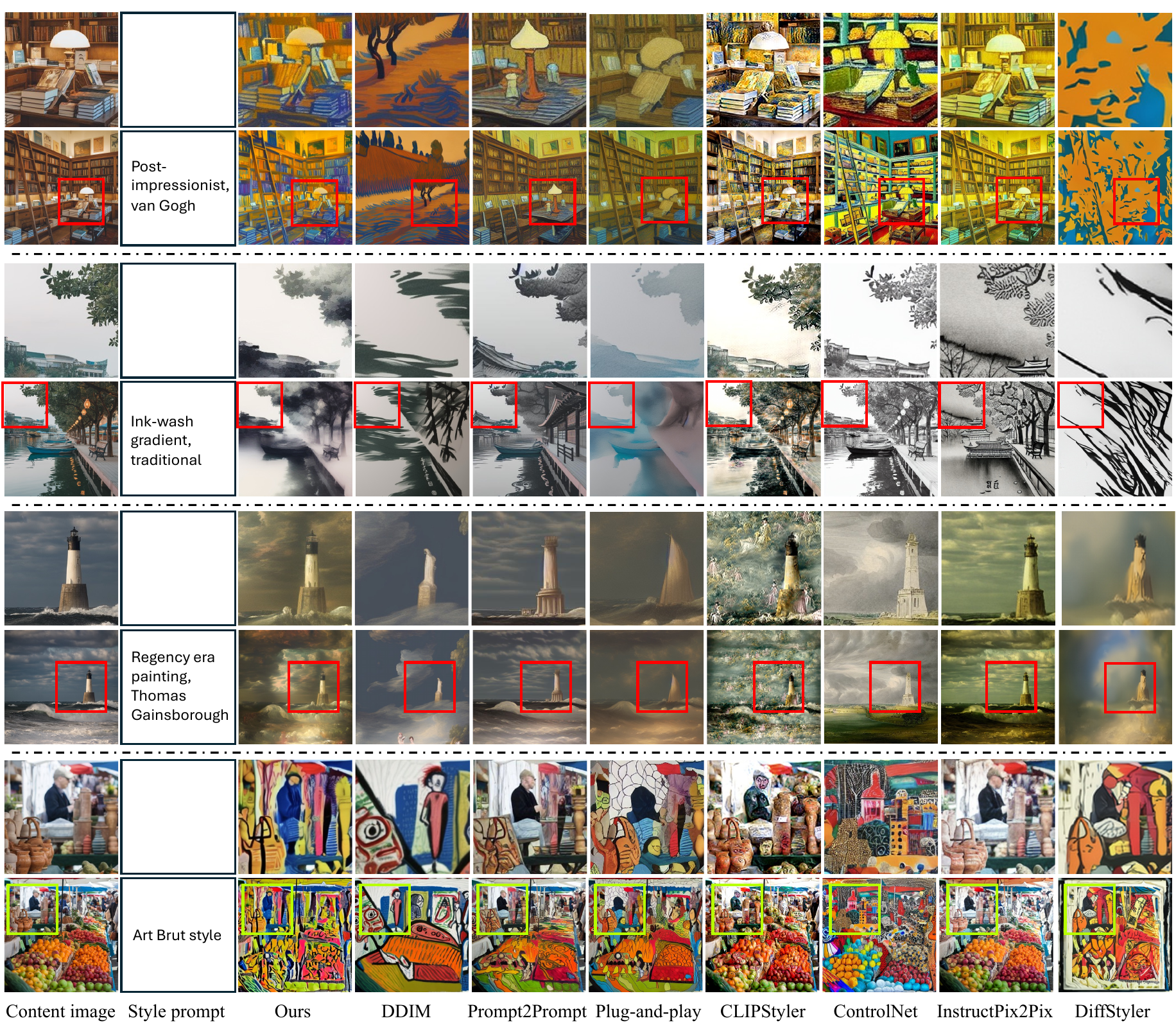}
    \caption{Extended comparison with existing text-driven stylization and manipulation methods.}
    \label{fig:supp_qual_cmp_1}
\end{figure*}

\begin{figure*}[!htbp]
    \centering
    \includegraphics[width=\linewidth]{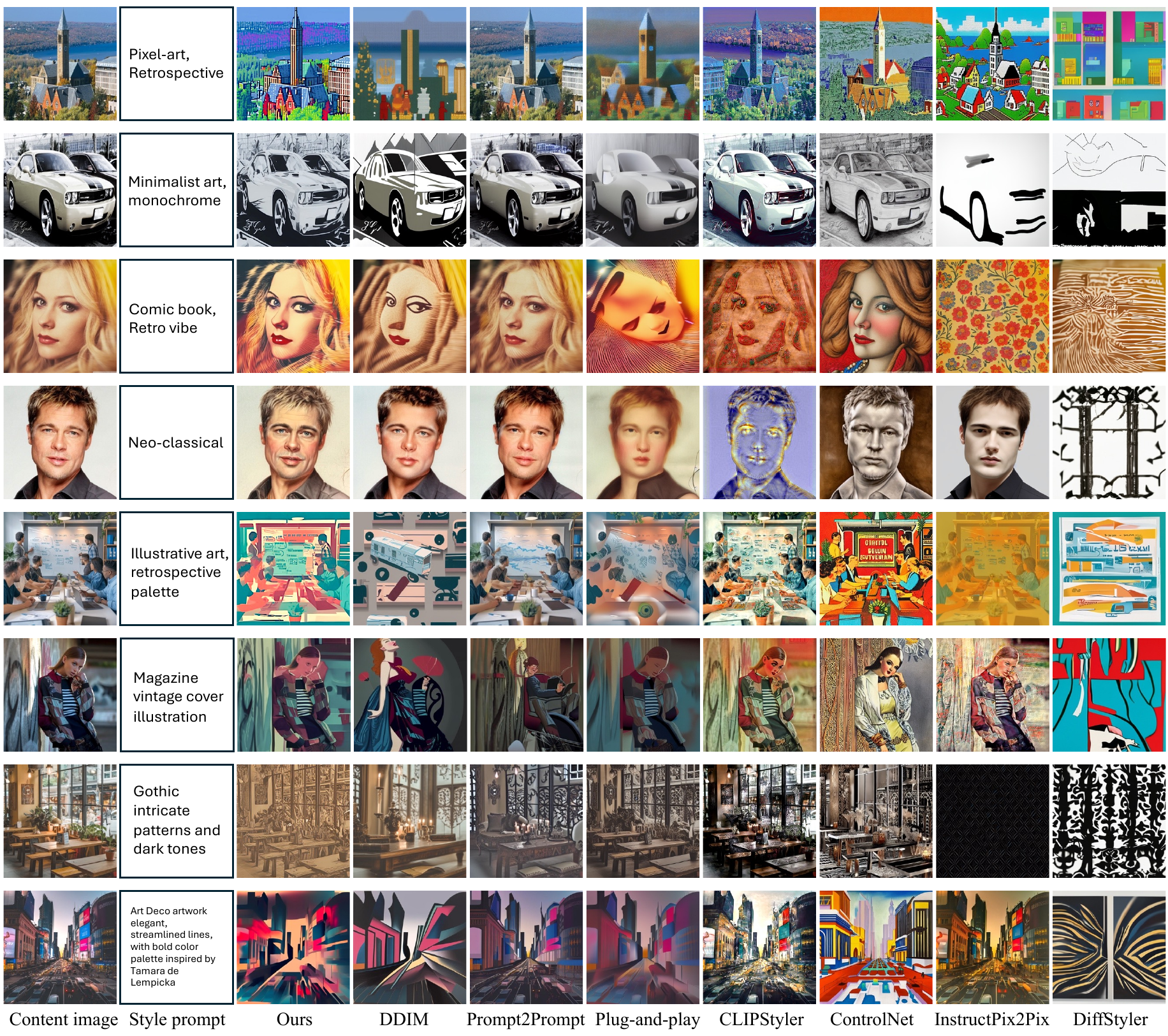}
    \caption{Extended comparison with existing text-driven stylization and manipulation methods.}
    \label{fig:supp_qual_cmp_2}
\end{figure*}

\textbf{Additional qualitative comparisons on stylization.}
We show an extended comparison with the previous \textit{reference-based} method in Fig.~\ref{fig:supp_ref_cmp_1}. More qualitative comparisons with existing \textit{text-driven} image stylization and manipulation methods can be found in Fig.~\ref{fig:supp_qual_cmp_1} and Fig.~\ref{fig:supp_qual_cmp_2}.

\begin{figure*}[!htbp]
    \centering
    \includegraphics[width=\linewidth]{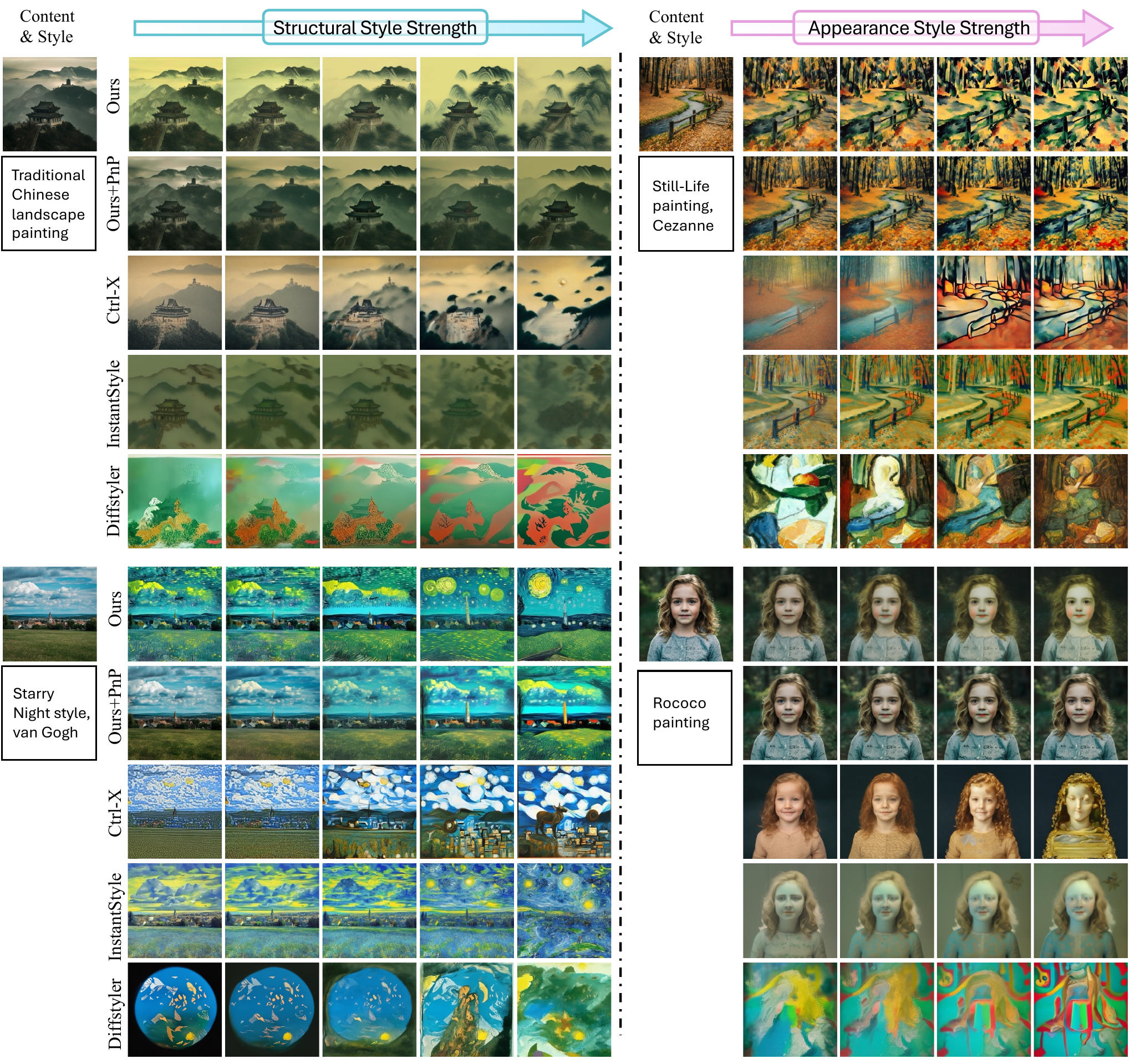}
    \caption{Extended comparison on fine-grained structural and appearance-based style control}
    \label{fig:supp_control_cmp_1}
\end{figure*}

\begin{figure*}[!htbp]
    \centering
    \includegraphics[width=\linewidth]{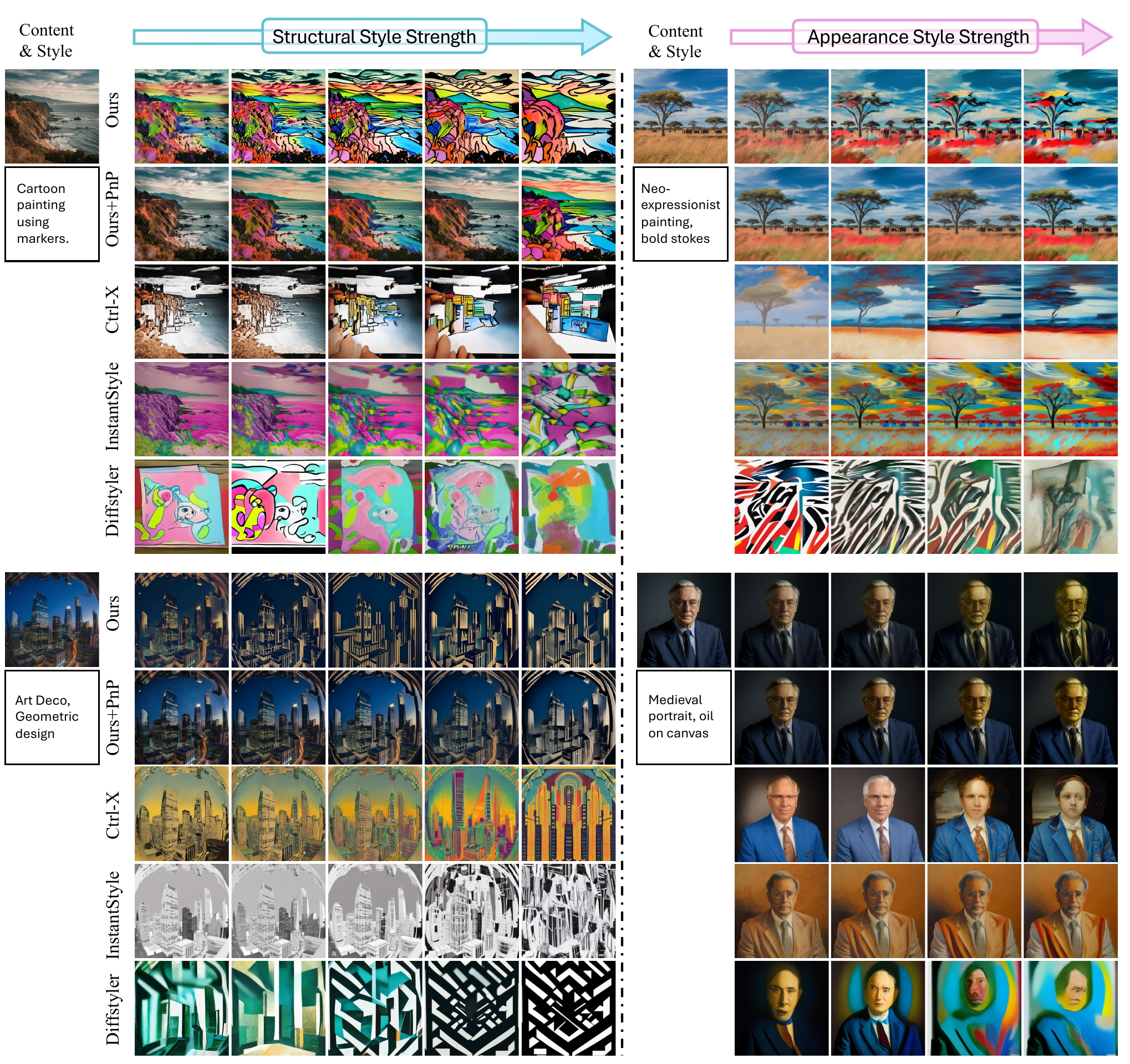}
    \caption{Extended comparison on fine-grained structural and appearance-based style control}
    \label{fig:supp_control_cmp_2}
\end{figure*}

\textbf{Additional comparisons on fine-grained control.} We provide additional comparisons with other control methods in Fig.~\ref{fig:supp_control_cmp_1} and Fig.~\ref{fig:supp_control_cmp_2}. These results demonstrate the advantage of DiffArtist in providing disentangled structural and appearance-level style control. In particular, the Ctrl-X, as an image editing method, produces less visually pleasing results when applied to image stylization. This is because they have a different definition of appearance and structure for editing real photos.

\section{On the Structure and Appearance Entanglement in Diffusion Process}\label{append: detail}

\subsection{Theoretical analysis} 
We now explore how the factorization of structure and appearance factorization, defined in Eq.~\ref{eqn:entangle}, interact and evolve throughout the denoising trajectory $p_\theta(\mathbf{x}_0,\mathbf{x}_1,\ldots,\mathbf{x}_{T-1}, |\mathbf{x}_T)$.
Specifically, suppose the content image $I_c$ is inverted into $\mathbf{x}_{1:T}$, inversion-based stylization starts from an intermediate step $\mathbf{x}_{\tau},\tau < T$ for DDIM denoising process. By rearranging Eqn.~\ref{eqn:ddim_reverse}, we obtain:

\begin{equation}
    \begin{aligned}
        \mathbf{x}_{t-1} &= \mathcal{A}_t \mathbf{x}_{t} + \mathcal{B}_t \epsilon_\theta(\mathbf{x}_t,t;y), \quad
        \mathcal{A}_t := \frac{\sqrt{\alpha_{t-1}}}{\sqrt{\alpha_{t}}} \\
        \mathcal{B}_t &:= \sqrt{1-\alpha_{t-1}} - \frac{\sqrt{\alpha_{t-1}(1-\alpha_{t-1})}}{\sqrt{\alpha_{t}}}
        \label{eqn: ddim_reverse_re}
    \end{aligned}
\end{equation}

Based on above formulation, the full stylization process could be expressed as:

\begin{equation}
\begin{aligned}
\hat{\mathbf{x}}_0 = \prod_{j=1}^{T} \mathcal{A}_j \cdot \mathbf{x}_T &
+ \underbrace{\sum_{k=\tau+1}^{T} \left[ \mathcal{B}_k \prod_{j=\tau+1}^{k-1} \mathcal{A}_j \right] \epsilon'(\mathbf{x}_{k-1}, k)}_{\text{preserve original structure and appearance}}\\
&
+ \underbrace{\sum_{k=1}^{\tau} \left[ \mathcal{B}_k \prod_{j=1}^{k-1} \mathcal{A}_j \right] \epsilon_\theta(\mathbf{x}_{k-1}, k; y),}_{\text{generate new structure and appearance}}
\label{eqn: linear}
\end{aligned}
\end{equation}
where $\epsilon'$ denotes an ideal denoiser that perfectly models the transition distribution $q(\mathbf{x}_{t-1} \mid \mathbf{x}_t)$. In Eq.~\ref{eqn: linear}, the conceptual denoising term from $\mathbf{x}_T$ to $\mathbf{x}_\tau$ preserves the structure and appearance in $I_c$. The stylization trajectory from $\mathbf{x}_\tau$ to $\mathbf{x}_t$ is what is actually computed, which introduces the desired appearance based on the prompt $y$ but may also lead to \textit{uncontrolled} structure generation. In this paradigm, preserving original appearance and generating new structure are undesirable side effects.
Intuitively, a larger $\tau$ leads to a shorter trajectory of $\mathbf{x}_T\to\mathbf{x}_\tau$, resulting in a stronger appearance in $\hat{\mathbf{x}}_0$ with weaker structure preservation (uncontrolled structure stylization), while a lower $\tau$ sacrifices stylization strength for stronger structure preservation. In other words, one can not arbitrarily control the strength of appearance and structure without affecting the other.

By combining Eqn.~\ref{eqn:entangle} and Eqn.~\ref{eqn: linear}, we can quantitatively assess the strength of content preservation and stylization in the diffusion process under a particular noise schedule $\alpha_{1:T}$. Specifically, we further assume the SNR of structure and appearance has a linear relationship with that of the $\mathbf{x}_t$ for $t$:
\begin{equation}
    \operatorname{SNR}(\mathbf{z}_t^c) \propto \frac{\alpha_t}{1-\alpha_t};     \operatorname{SNR}(\mathbf{z}_t^s) \propto \frac{\alpha_t}{1-\alpha_t}
\end{equation}

\subsection{Simulation}To derive the theoretical trends of structure and appearance strength during the diffusion process, we introduce an additional assumption. Specifically, we assume that the relative significance of each unweighted denoising step $\epsilon(\mathbf{x}_t,t;y)$ on the final stylized image remains consistent across different timesteps for both structure and appearance. In other words, we assume the SNR of structure and appearance has a linear relationship with that of the $\mathbf{x}_t$ at time $t$ as characterized by the noise schedule $\alpha_{1:T}$:
\begin{equation}
    \operatorname{SNR}(\mathbf{z}_t^c) \propto \frac{\alpha_t}{1-\alpha_t};     \operatorname{SNR}(\mathbf{z}_t^s) \propto \frac{\alpha_t}{1-\alpha_t},
\end{equation}
It is important to note that we do not assume that the relative proportions of structure and appearance are identical at each denoising step.

With the above assumption, the effect of varying $\tau$ on the structure and appearance of the final stylized image could be derived in closed form. In practice, we use the following code to calculate iteratively:

\begin{lstlisting}[language=Python]
def cum_score(low_t, hi_t, alphas):
    res = 0
    for k in range(low_t + 1, hi_t):
        for j in range(low_t + 1, k - 1):
          res += A_t(j, alphas) * B_t(k, alphas)
    return res

struct_scores = []
appear_scores = []

for tau in range(0, 50):
    crt_struct = cum_score(50 - tau, 50, sampled_alphas)
    crt_appear = cum_score(0, tau, sampled_alphas)
    struct.append(crt_struct)
    appear_scores.append(crt_appear)
\end{lstlisting}

\subsection{Empirical Result} Due to the inherent inaccuracy of DDIM inversion, the estimation of $\mathbf{x}_\tau$ may be imperfect, resulting in unintended modifications in the final sampled image even if no style prompt $y$ is used. To address this issue, we adopt an alternative strategy by randomly sampling 500 Gaussian noise as the $\mathbf{x}_T$ of content, which are paired with 500 content prompts. We treat the images denoised using content prompts for $\tau = T$ steps as the content image, which simulates a perfect inversion technique. To stylize an image, we first denoise the $\mathbf{x}_T$ with the content prompt for $\tau$ steps, which is subsequently denoised with the style prompt for $T-\tau$ steps. The LPIPS between the stylized image and the content image is used as the empirical structural strength. In contrast, the CLIP Deception score (correct classification rate among a set of styles) is used as the empirical appearance strength. The following 10 style prompts are used:
\begin{itemize}
    \item         "watercolor style"
        \item            "fauvism style"
        \item            "pencil sketch style"
        \item            "pointillism style"
         \item           "art deco style"
          \item          "impressionism style"
          \item          "surrealism style"
        \item            "pop art style"
        \item           "cubism style"
         \item           "abstract expressionism style"
\end{itemize}

The results for both simulation and empirical results are in Fig.~\ref{fig:trajectory}. The result shows a good fit, and it turns out that the structure modification appears to be linear, with the stylization strength being quadratic with respect to $\tau$. Moreover, this result further evidenced the issue of S-A entanglement in the diffusion process.
\section{Details on MLLM-based metrics}
\label{append: MLLM}
\subsection{Implementation Details}
The stylized images, style prompt, and the instruction prompts are fed to MLLM for inference. We compose stylized images as a grid image with numbers at the top-left corner. The full prompt template for structure and appearance score is available in Tab.~\ref{tab:prompt}.



\subsection{Correlation with Human Preference}

\begin{figure*}[!htbp]
    \centering
    \includegraphics[width=\linewidth]{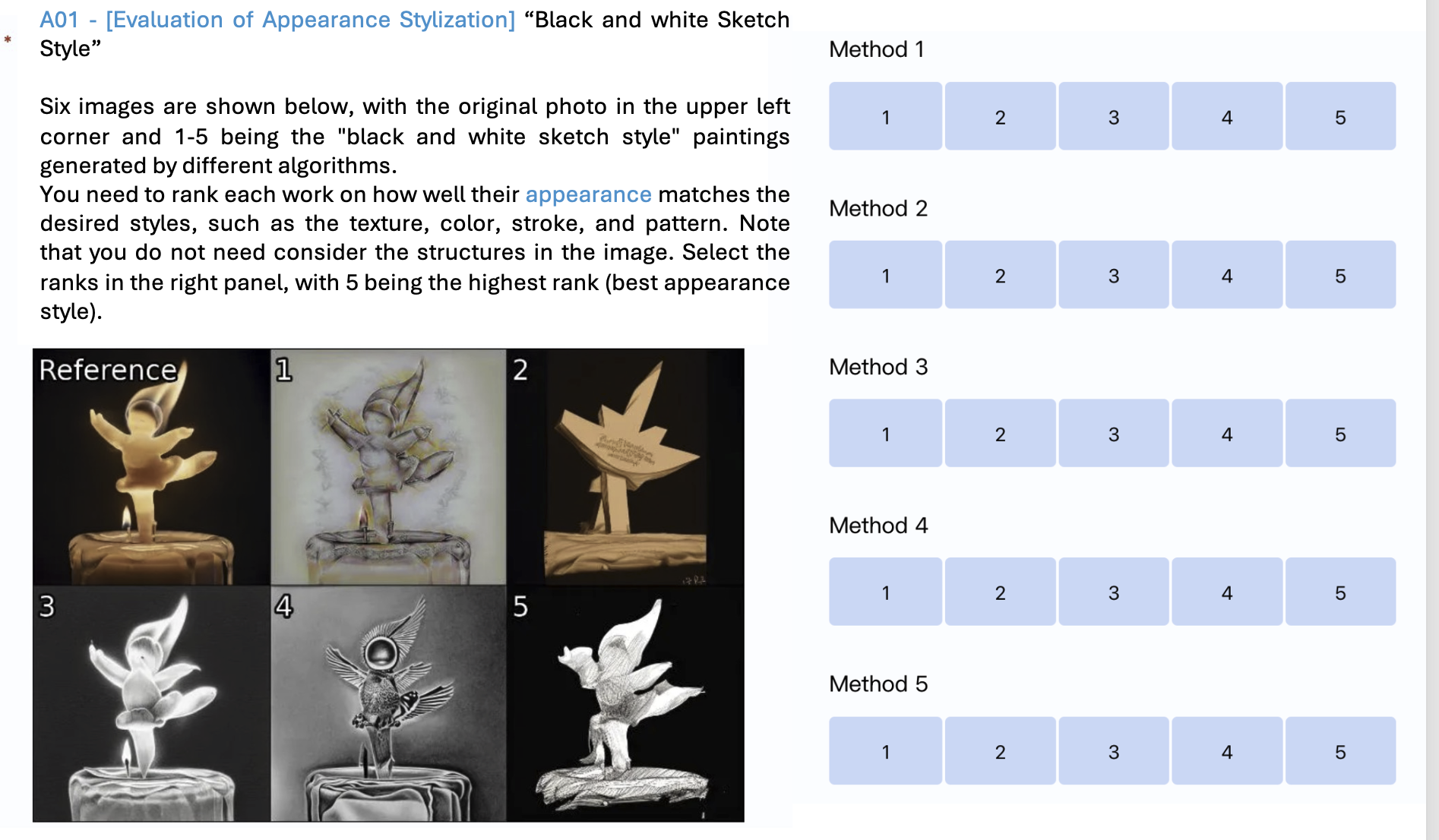}
    \caption{Example user interface in collecting human preference. The system will prevent user from selecting the same ranking.}
    \label{fig:ui}
\end{figure*}

\textbf{Human Question Collection} We distributed the questionnaire on a crowd-sourcing platform, where each participant was required to complete up to 20 randomly sampled ranking tasks. An example of the user interface is provided in Fig.~\ref{fig:ui}. A total of 200 participants took part in this study. To ensure the validity of the responses, we included attention-check questions. If a participant answers an attention-check question incorrectly, all of their responses will be marked invalid. Responses that are made with less than 20 seconds are also removed.

\begin{table*}[!htbp]
\caption{Prompt templates for MLLM-based metrics. [IMG], [STYLE] and [NUM\_METHOD] is the placeholder for combined image, style and number of methods, respectively.}
\centering
\renewcommand{\arraystretch}{1.5} 
\setlength{\tabcolsep}{10pt} 
\begin{tabular}{|p{0.45\textwidth}|p{0.45\textwidth}|}
\hline
\textbf{Structure Score} & \textbf{Appearance Score} \\ \hline
            "[IMAGE] A content (source) image (top left) and [NUM\_METHOD] stylized images in the style of [STYLE] are placed as a grid. "
            "The stylized images are indexed from left to right, and from top to bottom. "
            "Compare, analyze and discriminately rank the fidelity to which the structure described in the style of [STYLE] is transferred to the source image."
            "You should focus on the fidelity of structure-related style component only, such as the lines, shapes, geometry, layout, and perspective.  You should not consider the style related to appearance (e.g., texture, color, stroke, and pattern). You should also consider how the structure of [STYLE] is integrated with the source image."
            "Stylized image that has (1) limited style strength, (2) structure that is mis-aligned with the style, or (3) significant artifacts and distortions of the semantic integrity  (e.g., the original object and scene become unrecognizable) unless the distortion is explicitly intended by the style of  [STYLE], and (4) un-harmonious integration with the source image should be considered of in lower rank.   In other words, if a stylized image is not an artistically meaningful painting of the source image in target style, then it should be rated lower. Images that harmoniously integrate the structure of [STYLE] with the source image should be rated higher."
            "Rank the [NUM\_METHOD] images in ascending order from 1 to [NUM\_METHOD], where the highest rank of [NUM\_METHOD] means the best structural fidelity. No images shall have the same ranking. "
            "As an expert in art, return your thinking in short (what structure is desired, and how the ranking is decided for each image in short), and ranks for each image id in a Python Dict, ['thinking':str, 'rank':List[[NUM\_METHOD]]].  Do not include any other string in your response." &
            "[IMAGE]  A content (source) image (top left) and [NUM\_METHOD] stylized images in the style of [STYLE] are placed as a grid. "
            "The stylized images are indexed from left to right, and from top to bottom. "
            "Compare, analyze and discriminately rank the fidelity to which the appearance described in the style of [STYLE] is transferred and to the source image."
            "You should focus on the fidelity of appearance-related art style component only, such as the texture, color, stroke, and pattern. Note that it does not simply means color pallette and saturation. You should not consider the style related to structure (e.g., lines, shapes, geometry, layout, and perspective), unless the original scene become unrecognizable.  You should also consider how the appearance of [STYLE] is integrated with the source image."
            "Stylized image that has (1) limited style strength, (2) visual appearance that is mis-aligned with the style, (3) significant artifacts and distortions of the semantic integrity (e.g., the original object and scene become unrecognizable) unless the distortion is explicitly intended by the style of  [STYLE] and (4) un-harmonious integration with the source image should be considered of in lower rank. In other words, if a stylized image is not an artistically meaningful painting of the source image in target style, then it should be rated lower. Images that harmoniously integrate texture, color, stroke, and pattern should be rated higher."
            "Rank the [NUM\_METHOD] images in ascending order from 1 to [NUM\_METHOD], where the highest rank of [NUM\_METHOD] means the best appearance fidelity. No images shall have the same ranking."
            "As an expert in art, return your thinking (what appearance is desired, and how the ranking is decided for each image in short), and ranks for each image id in a Python Dict, ['thinking':str, 'rank':List[[NUM\_METHOD]]]. Do not include any other string in your response."
\\ \hline
\end{tabular}

\label{tab:prompt}
\end{table*}

\section{Additional Discussion and Analysis}
\label{append: analysis}

\subsection{Generalizability of DiffArtist}
To demonstrate the generalizability of DiffArtist across different U-Net-based diffusion architectures, we implement our method on Playground v2\footnote{\url{https://huggingface.co/playgroundai/playground-v2-1024px-aesthetic}}, which utilizes the SDXL architecture, distinct from Stable Diffusion 2.1. Several results are provided in Fig.~\ref{fig:playground}. These results validate that DiffArtist serves as a versatile control method applicable to various U-Net-based diffusion models, regardless of their underlying architectural differences.

\subsection{Additional Ablations on S2A Injection}\label{sec:ablate_S2A}

\begin{figure*}[!htbp]
    \centering
    \includegraphics[width=0.8\linewidth]{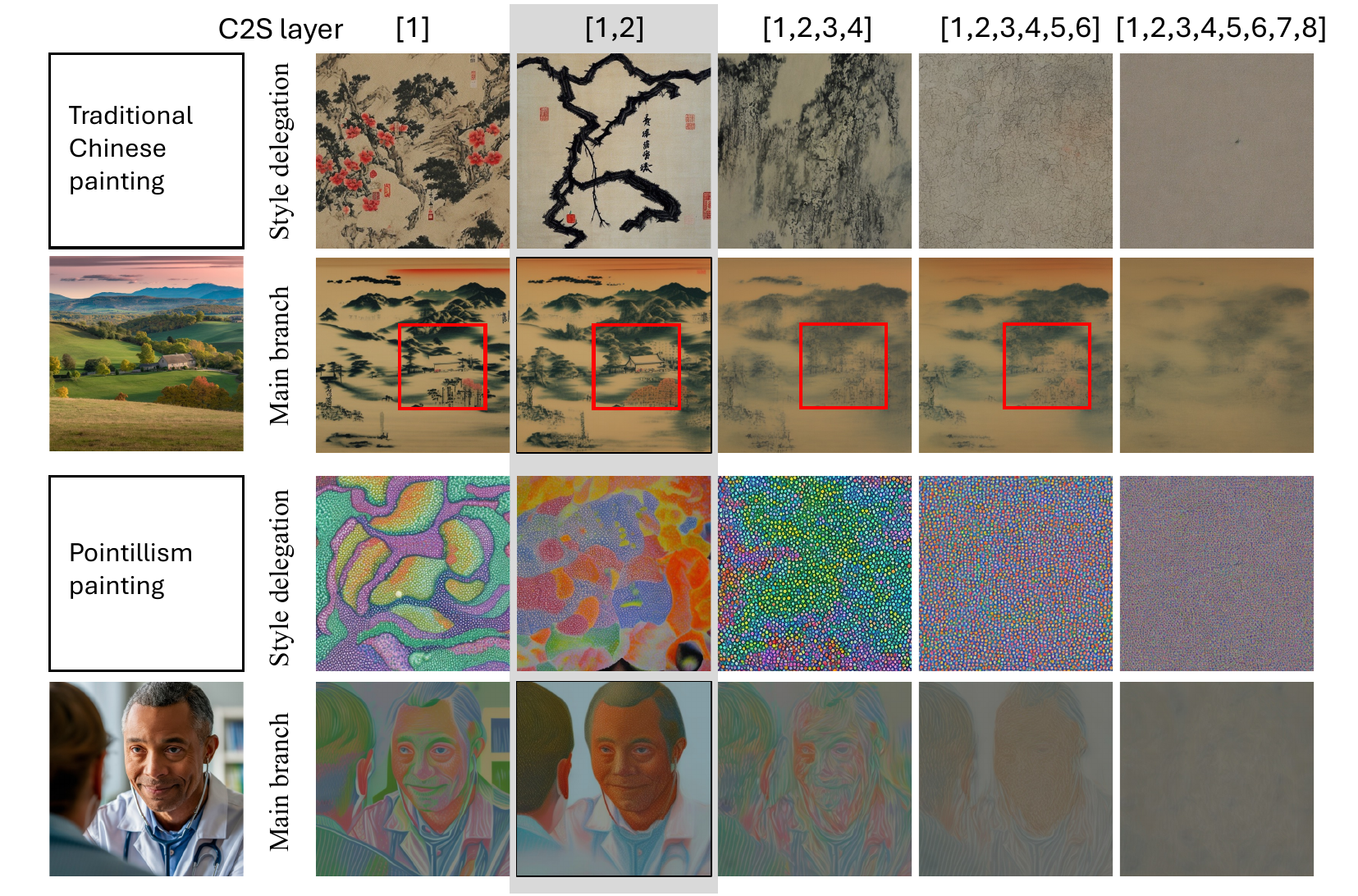}
    \caption{\textbf{Ablation Study on S2A Layers \( S_{S2A} \).} Increasing the number of S2A layers compels the appearance delegation to generate higher-frequency style features (stokes, points) while diminishing low-frequency tonal components (large color fields). Empirically, our default configuration \([1, 2]\) strikes an optimal balance between enhancing style detail and preserving essential content structure.}
    \label{fig:S2A_layers}
\end{figure*}

In this section, we provide additional ablations to study the effect of proposed S2A design. 

\textbf{Ablation on S2A layers $S_{S2A}$}. We ablate the number of injection layer used in the S2A injection (i.e., $S_{S2A}$). As illustrated in Fig.~\ref{fig:S2A_layers}, the S2A layers influence the frequency bands of style delegation. Incorporating only early layers (e.g., \verb|[1, 2]|) focuses on generating low-frequency style features such as tones and small objects, while adding more layers facilitates the creation of high-frequency style details like stroke shapes. Empirically, we set the S2A injection layers to \verb|[1, 2]| by default, as using additional layers typically results in blurriness in the stylized outputs.

\subsection{Failure Case}
In our experiment, we identify a rare ($<$1\%) and special failure case in the proposed methods. Specifically, for certain content image, its stylization result will consistently contains black and white chessboard-pattern artifacts. We provide one example in Fig.~\ref{fig:fail}.
    
\end{appendices}

\end{document}